\documentclass[10pt,twocolumn,letterpaper]{article}

\usepackage[pagenumbers]{cvpr}

\usepackage{colortbl}
\definecolor{eraserow}{HTML}{F6DFE8}
\definecolor{erasemrow}{HTML}{EFE0F3}
\usepackage{multirow}
\usepackage{pifont}

\usepackage{amsthm}

\newtheoremstyle{paperplain}{\topsep}{\topsep}{\itshape}{}{\bfseries}{.}{.5em}
{\mbox{\thmname{#1}\thmnumber{ #2}\thmnote{ {\normalfont(#3)}}}}
\newtheoremstyle{paperdefinition}{\topsep}{\topsep}{\normalfont}{}{\bfseries}{.}{.5em}
{\mbox{\thmname{#1}\thmnumber{ #2}\thmnote{ {\normalfont(#3)}}}}
\theoremstyle{paperplain}

\newtheorem{theorem}{Theorem}[section]
\newtheorem{proposition}[theorem]{Proposition}
\theoremstyle{paperdefinition}
\newtheorem{definition}[theorem]{Definition}
\DeclareMathOperator*{\argmax}{arg\,max}
\let\ERASEsum\sum
\renewcommand{\sum}{\ERASEsum\limits}
\let\ERASEmin\min
\renewcommand{\min}{\ERASEmin\limits}
\newcommand{\R}{\mathbb{R}}
\newcommand{\cX}{\mathcal{X}}
\newcommand{\cY}{\mathcal{Y}}
\newcommand{\cCf}{\mathcal{C}_f}
\newcommand{\vx}{\mathbf{x}}
\newcommand{\vz}{\mathbf{z}}
\newcommand{\vw}{\mathbf{w}}
\newcommand{\vb}{\mathbf{b}}
\newcommand{\mW}{\mathbf{W}}
\newcommand{\HMt}{\mathrm{HM}_t}
\usepackage{algorithm}
\usepackage{algorithmic}

\newcommand{\mP}{\mathbf{P}}
\newcommand{\vmu}{\boldsymbol{\mu}}
\newcommand{\vt}{\mathbf{t}}
\newcommand{\vu}{\mathbf{u}}

\definecolor{cvprblue}{rgb}{0.21,0.49,0.74}
\usepackage[pagebackref,breaklinks,colorlinks,allcolors=cvprblue]{hyperref}

\title{Forgotten but Recognizable: Erasing Classes from Released Features}

\author{Junhao Cai \qquad Dohun Kim \qquad Changhee Joo\thanks{Corresponding author.}\\
Korea University, Seoul, Republic of Korea\\
{\tt\small \{junhochae, dohunkim, changhee\}@korea.ac.kr}
}

\usepackage{flafter}

\newcommand{\pmstd}[1]{{\tiny\,\textpm\,#1}}
\newcommand{\hz}{\phantom{0}}
\newcommand{\erasem}{\mbox{ERASE-M}}
\newcommand{\resc}{\mbox{R-ESC}}
\newcommand{\pourp}{\mbox{POUR-P}}

\begin{document}
\maketitle
\begin{abstract}
Class unlearning removes a set of classes from a trained model, with retraining on the remaining classes as the reference.
When features are released, a linear probe trained on them still recognizes the forget classes in retrained models, more so for more accurate models.
This implies that deleting the data of a class does not delete the class from the features.
An accurate model already knows where the features of its remaining classes lie.
\textbf{ERA}sure by \textbf{SE}lf Knowledge (ERASE) therefore predicts the location and shape of each remaining class from the frozen model, without remaining-class samples, and moves each forget class onto them by Gaussian optimal transport.
ERASE with class \textbf{M}oments (\erasem) instead takes them from moments stored for every class before deletion, which also keeps detectors near chance.
Across classification, segmentation, and detection with diverse networks and backbones, new classifiers rarely recognize the forget classes, and both methods outperform baselines on most benchmarks.\par
\end{abstract}

\section{Introduction}
\begin{figure}[t]
\centering
\includegraphics[width=\columnwidth]{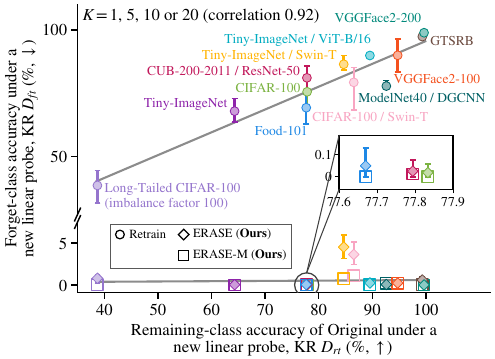}
\caption{\textbf{Recovery after data exclusion.} KR forget-class accuracy $D_{ft}$ of Retrain, ERASE, and \erasem{} against KR remaining-class accuracy $D_{rt}$ of Original, colored by dataset~\cite{bossard2014food,wah2011caltech,cui2019class} and network~\cite{dosovitskiy2020image,liu2021swin} (\mbox{ResNet-18} unless named).
Points average three forget-set sizes $K$ ($K\leq10$ on GTSRB~\cite{stallkamp2011german} and ModelNet40~\cite{wu20153d}), and bars span them.
Least-squares lines over the 36 settings fit Retrain (upper, correlation in title) and ERASE and \erasem{} (lower).}
\label{fig:teaser}
\end{figure}

Class unlearning aims to remove a set of forget classes from a trained model while keeping its accuracy on the remaining classes~\cite{golatkar2020eternal,chundawat2023can,zhou2025decoupled}.
A right to erasure motivates this removal~\cite{eu2016general}, with retraining without the forget classes as the reference~\cite{cao2015towards,bourtoule2021machine}.
Most methods update the trained weights to approach this reference and evaluate the forget-class accuracy of the deployed classifier.

A model consists of a feature extractor and a head that maps features to class scores.
When a service releases the extracted features of its unlearned model, a recipient can train its own classifier on them, and this classifier may still recognize the forget classes.
A face service that deletes a person on request, for instance, keeps releasing face features from which a recipient with photos of that person can train a classifier to recognize them.
Knowledge deletion addresses this risk and asks that no such classifier recognize the forget classes either, which the Knowledge Retention (KR) evaluation of~\cite{lee2025esc} tests by training a linear classifier on the features.
Retraining does not meet this requirement, since this linear classifier still recognizes the forget classes in the features of retrained models, and more so for more accurate models (\cref{fig:teaser}).
Classifiers in prior evaluations also recover these classes from retrained models~\cite{jeon2026r,gao2026illusion}.
Accurate features separate even classes that the model never saw, so deleting the data of a class does not delete the class from the features.
Retraining thus meets the usual goal of removing the influence of the forget data but not that of knowledge deletion, and we judge deletion on the released features.

The features themselves must therefore change.
Prior feature-level methods keep the weights and edit the features that the deployed model serves, by projecting them~\cite{lee2025esc,le2026pour,fang2026machine} or by moving forget-class means~\cite{jeon2026r}.
A newly trained classifier still tells their edited forget-class features apart from real remaining-class features, so changing the features is not enough.
\emph{What should a forget class look like in the features?}

Each of its features should follow the distribution of a remaining class.
An edited feature then looks like a real feature of that class, so a newly trained classifier cannot tell the two apart.
An edit thus has to match both the location and shape of this distribution.
Estimating them seems to need remaining-class samples, which the service may no longer hold.
An accurate model already knows where the features of its remaining classes lie, so \textbf{ERA}sure by \textbf{SE}lf Knowledge (ERASE) predicts this distribution from the frozen model.
The head weights and two class-balanced moments stored before the deletion request give its location, and the forget-class features give its shape.
Gaussian optimal transport then moves the forget classes onto real remaining classes, not onto a pooled or barycentric target as in concept erasure.
Among the maps that match this location and shape, it moves the features the least and is invertible, so it keeps the differences new tasks need.
Across the models of \cref{fig:teaser}, the recovery after ERASE stays near zero, so using what the model knows helps most where retraining fails most.

The contributions of our work are as follows:
\begin{itemize}
\item We show that a linear probe recognizes the forget classes in models retrained only on the remaining classes, more so for more accurate models.
We bound Bayes recovery and detector AUC by the divergence from a remaining class, which vanishes only if the location and the shape both match.
Projections, mean shifts, and pooled targets miss this match and stay detectable.
\item We propose ERASE, which turns the accuracy that defeats retraining into the means of deletion: it predicts remaining-class distributions from the frozen model and transports the forget classes onto them.
Its extension \erasem{} takes them from class moments stored before deletion, which undetectability needs.
\item We evaluate ERASE and \erasem{} on classification, segmentation, and detection with diverse networks and frozen backbones.
Both keep forget-class accuracy near zero regardless of model accuracy, outperform baselines on most benchmarks, and handle a deletion request orders of magnitude faster than retraining a trained network.
\erasem{} also keeps detectors near chance.
\end{itemize}

\section{Related Work}
Parameter-based methods fine-tune on the remaining data~\cite{golatkar2020eternal}, ascend the loss of the forget data~\cite{golatkar2020eternal}, relabel it randomly~\cite{graves2021amnesiac}, update salient weights~\cite{fan2024salun}, or distill from a randomly initialized~\cite{chundawat2023can} or masked teacher~\cite{zhou2025decoupled}.
Other methods take spectral steps~\cite{choi2026spectral} or min-max steps with retain-neutral surrogates~\cite{cai2026retain}.
Loss ascent on clean data also detects backdoored clients in federated learning~\cite{kim2026data}.
ELM~\cite{gandikota2025erasing} fine-tunes a language model and trains linear probes on its internal representations to detect the erased knowledge.

Feature-level methods prune forget-class feature directions~\cite{lee2025esc}, project out forget-class directions of the classifier~\cite{le2026pour}, learn a projection from forget and remaining statistics~\cite{fang2026machine}, or condition a projection on the input~\cite{cai2026scope}.
A projection leaves forget-class information in the directions it keeps or removes remaining-class information with those it drops.
\resc~\cite{jeon2026r} instead moves forget-class means onto remaining ones at random, so it matches only the location.

Concept erasure edits representations so that a classifier cannot use a concept.
LEACE~\cite{belrose2306leace} equalizes the class means by a least-squares affine map, QLEACE~\cite{quirke2025slowing} also equalizes their covariances, and fairness repair moves groups toward a common barycenter~\cite{feldman2015certifying,gordaliza2019obtaining}.
These maps move every class or group and need labeled samples of each.
Gaussian optimal transport maps one class onto another~\cite{singh2024representation}.
ERASE transports only the features it selects, toward remaining classes whose location and shape it predicts from the frozen model, and \erasem{} takes these from class moments stored before deletion.

\section{Problem Formulation and Analysis}
\subsection{Notation}
Let $\cX$ be the input space and $\cY=\{1,\ldots,C\}$ the class set.
The frozen model has an extractor $g:\cX\to\R^d$ and an affine head $h:\R^d\to\R^C$ with weights $\mW\in\R^{C\times d}$, rows $\vw_j^\top$, and bias $\vb$.
For $\vz=g(\vx)$, the logits are $h(\vz)=\mW\vz+\vb$.
Write $p_j(\vz)=\mathrm{softmax}(h(\vz))_j$ and $F(\vx)=\argmax\nolimits_{j\in\cY}p_j(g(\vx))$, where all maximizers break ties by class order.
The orthogonal projector onto the row space of $\mW$ is $\mP$.

The training samples form an indexed collection $\mathcal S=((\vx_i,y_i))_{i=1}^{N}$.
For each class $c \in \cY$, let $\mathcal I_c=\{i:y_i=c\}$, $N_c=|\mathcal I_c|$, and $\vmu_c=\frac{1}{N_c}\sum\nolimits_{i\in\mathcal I_c}g(\vx_i)$, and assume $N_c\geq2$.
The forget classes are $\cCf\subset\cY$ with $0<K=|\cCf|<C$, the remaining classes are $\mathcal C_r=\cY\setminus\cCf$.
We use natural logarithms and $\mathrm{KL}$ for relative entropy, and the proofs are in the appendix.

\subsection{Edits and Recovery}
A hosted feature service keeps $g$ and $h$ private and unchanged, so knowledge deletion is judged on the features that it releases, all that recipients or downstream nodes~\cite{cai2026hcf,cai2024toward,bae2026rxc} see.
Each query carries an input $\vx\in\cX$ and a key $q\in\mathcal Q$, such as the encoded bytes of the input, which lets the service recognize repeated queries.
Recipients can train new classifiers on these features with labels they hold, including forget-class labels, while the service sees no query label.
The service needs knowledge deletion when it must stop releasing features that reveal a class, such as a person who objects to identification or a category its policy prohibits.
\begin{definition}[Feature edit]
Given the stored state of the service (\cref{sec:method}), a feature edit $r:\cX\times\mathcal Q\to\R^d$ supplies $r(\vx,q)$ in place of $g(\vx)$ to the recipient and the deployed head $h$.
\end{definition}
\noindent We write $\tilde\vz=r(\vx,q)$ for the received feature and suppress $q$ when each input has a fixed key.

\begin{definition}[Unrecognizable and undetectable classes]\label[definition]{def:recovery}
A probe $\varphi:\R^d\to\cY$ is trained on the edited training features $\tilde\vz_i$ and labels $y_i$. A test input $\vx$ with label $y$ is drawn from a distribution that puts positive probability on both $\cCf$ and $\mathcal C_r$, and $\tilde \vz = r(\vx)$.
The \emph{recovery} of $\varphi$ is the forget-class test accuracy $\Pr[\varphi(\tilde\vz)=y\mid y\in\cCf]$.
A detector is trained on these features with labels $\mathbb I\{y_i\in\cCf\}$.
For a tolerance $\gamma\geq0$, the forget classes are \emph{unrecognizable} to a family of probes if every probe of the family trained this way has recovery at most $\gamma$.
They are \emph{undetectable} by a family of detectors if the test AUC of every detector in the family is within $\gamma$ of $\frac{1}{2}$.
\end{definition}
Knowledge deletion asks that the forget classes be unrecognizable to the linear probe of KR~\cite{lee2025esc}.

The edit must also preserve utility.
The deployed head should keep its accuracy on the remaining classes, and the released features should still serve the uses that the service declares.
These uses include remaining-class classification and new tasks whose labels lie outside $\cY$, such as new identities for a face model (\cref{sec:exp-utility}).
We compare with two models.
Original is the model $(g,h)$ trained on all classes, and Retrain is trained only on the samples of $\mathcal C_r$, the usual reference for removing the influence of the forget samples.

\subsection{Features of Retrained Models}
\label{sec:observation}
We measure the recovery $D_{ft}$ of the KR linear probe on retrained features, over 66 settings of dataset, network, and number $K$ of forget classes (\cref{fig:teaser} and Appendix).
In \cref{fig:teaser}, $D_{ft}$ rises with the KR remaining-class accuracy of Original for each $K$ and with Normal $D_{rt}$ over retraining epochs (Appendix).
An MLP probe also recovers the forget classes, and both probes recover far less from a randomly initialized network, for example 4.61\% against 75.50\% for Retrain under KR on \mbox{CIFAR-100} with \mbox{ResNet-18} (Appendix).

\subsection{What an Edit Has to Keep and Match}
\label{sec:principle}
Retrained features stay recognizable (\cref{sec:observation}), so an edit must remove what probes use to recognize $\cCf$, while keeping what new tasks need.
Each edited forget-class input takes a remaining class, its target, as the deployed prediction.
The prediction keeps only this class, but the feature can also keep differences among inputs with the same target, which new tasks need.
A template edit maps all inputs of one target to one fixed vector, so on a new task no classifier of its features exceeds in expectation the rule predicting each target's most frequent label (proof in the appendix).
The bound also holds for draws from the target distribution, so accuracy above it shows information beyond the target.

For $c\in\cCf$ and $k\in\mathcal C_r$, let $Q_{ck}$ be the distribution of $\tilde\vz$ given $y=c$ and target $k$, and $P_k$ the distribution of $\tilde\vz$ given $y=k$.
Let $\varphi^\star$ be the Bayes classifier and $Q_{ck}(\varphi^\star=c)=Q_{ck}\big(\{\tilde\vz:\varphi^\star(\tilde\vz)=c\}\big)$ the probability that it recognizes a class-$c$ input with target $k$.
The next proposition bounds this probability.
\begin{proposition}[Recovery bound]\label[proposition]{prop:recovery}
Let class priors be equal and the deployed head assign each forget-class input to its remaining-class target.
Let target $k$ receive a share $\pi_{ck}\in(0,1]$ of class-$c$ inputs, equal to $\frac{1}{C-K}$ for uniform targets in $\mathcal C_r$.
If $\pi_{ck}<1$, the Bayes classifier $\varphi^\star$ on edited features satisfies
\begin{equation}
Q_{ck}(\varphi^\star=c)\leq\frac{\mathrm{KL}(Q_{ck}\,\|\,P_k)}{\pi_{ck}-1-\log\pi_{ck}}.
\end{equation}
If $Q_{ck}$ and $\pi_{ck}$ do not depend on $c\in\cCf$, every probe recovers at most $\frac{1}{K}$ of the forget-class inputs with target $k$.
\end{proposition}
For any feature distributions and fixed target shares, a smaller divergence gives a smaller bound.
The same divergence also bounds the AUC of detectors separating class-$c$ inputs with target $k$ from class-$k$ inputs, and their detection rate at every false-positive rate (Appendix).
Unlike the recovery bound, this bound does not depend on $\pi_{ck}$, so detectors stay at chance only when $Q_{ck}=P_k$.
For nondegenerate Gaussian laws, the divergence splits into location and shape terms, so an edit has to match $P_k$ in both.
Moving forget-class means onto remaining ones, as \resc{} does, matches only the location, and MLP detectors separate such edits (\cref{tab:capacity}).
For $K>1$, sharing one distribution and one target share within each target also keeps probes from telling the forget classes apart (\cref{prop:recovery}).

Matching $P_k$ requires its location and shape.
Stored class statistics give both.
Without them, the frozen model, the forget classes, and two class-balanced moments must predict both.
Such predictions cannot ensure undetectability, since class distributions sharing this information can admit detectors with AUC near one (proof in the appendix, following~\cite{radic2026forget}).
ERASE predicts both for knowledge deletion and \erasem{} stores them for undetectability.

\section{Methodology}
\label{sec:method}
ERASE edits each released feature with $g$ and $h$ fixed (\cref{alg:erase}).
It first replaces every feature $\vz$ by $\mP\vz$, which keeps the logits, since $\mW\mP=\mW$, and drops the directions that the head ignores, from which probes would otherwise recover the forget classes (\cref{sec:exp-ablation}). Then a gate selects the inputs predicted as forget classes and assigns each a target.
ERASE predicts the location and shape of each target distribution and transports each selected feature toward it while keeping its variation.
The prediction uses the frozen model, the forget-class training samples, and two class-balanced moments stored before a deletion request, which contain no individual sample.
For a query $\vx$ with key $q$, let $k$ be its target, $\vu$ its feature after transport, and $\beta^\star$ its final step along $\vw_k$, which sets the deployed prediction.
ERASE releases $\tilde\vz=r(\vx,q)=\vu+\beta^\star\vw_k$ (or scaled, \cref{sec:protect}).
Where ERASE predicts each target distribution, \erasem{} takes its mean and covariance from statistics stored for every class (\cref{sec:stored}).
Both fit the edit without remaining-class samples, and the service can delete the forget-class samples after computing the fitted quantities of \cref{alg:erase}.

\begin{algorithm}[t]
\caption{ERASE and \erasem}
\label{alg:erase}
\begin{algorithmic}[1]
\renewcommand{\algorithmicrequire}{\textbf{Input:}}
\renewcommand{\algorithmicensure}{\textbf{Output:}}
\REQUIRE {\raggedright \mbox{query $\vx$ with key $q$}, \mbox{target map $\psi^\star$}, \mbox{frozen model $(g,h)$}, \mbox{forget classes $\cCf$}, and the fitted \mbox{projector $\mP$} and \mbox{transport maps $\mathcal G_{cj}$}\par}
\item[\textbf{Parameters:}] {\raggedright \mbox{confidence parameter $\epsilon$}, \mbox{power $\zeta>0$}\par}
\ENSURE {\raggedright released feature $\tilde\vz=r(\vx,q)$\par}
\STATE $\vz\gets g(\vx)$
\STATE $\vu\gets\mP\vz$ \COMMENT{keeps the logits}
\STATE $k\gets F(\vx)$
\IF{$F(\vx)\in\cCf$}
\STATE $c\gets c(\vz)$ \COMMENT{source, \cref{sec:selection}}
\STATE $k\gets\psi^\star(\vx,q)$ \COMMENT{target, \cref{sec:selection}}
\STATE $\vu\gets\mathcal G_{ck}(\vu,q)$ \COMMENT{\cref{eq:move}}
\ENDIF
\STATE $\delta\gets\epsilon\bigl(1-p_k(\vu)\bigr)^{\zeta}$
\STATE $\beta^\star\gets\min\{\beta\geq0:p_k(\vu+\beta\vw_k)\geq1-\delta\}$
\STATE \textbf{return} $\tilde\vz=\vu+\beta^\star\vw_k$
\end{algorithmic}
\end{algorithm}

\subsection{Selecting Inputs and Targets}
\label{sec:selection}
The recovery bound assumes that the deployed head predicts a remaining class for every forget-class input (\cref{prop:recovery}), so a gate selects every input with $F(\vx)\in\cCf$.
Its source is $c(\vz)=\argmax\nolimits_{j\in\cCf}(\mW\vz)_j$ scored by the logit without the bias.
As in \resc~\cite{jeon2026r}, a selected input takes a random target, $\psi(q)$ for its key $q$ under the target map $\psi:\mathcal Q\to\mathcal C_r$.
When $q$ first appears, the service draws $\psi(q)$ from $\mathcal C_r$, uniformly for ERASE and in proportion to the training counts for \erasem, and keeps it, so repeated queries of one input receive the same feature.
The second part of the bound needs shares $\pi_{ck}$ that do not depend on $c$, so every forget class draws its targets from the same distribution.
Each forget class thus spreads over all remaining classes, so with two or more of them every share $\pi_{ck}$ of \cref{prop:recovery} is below one, and a smaller share gives a smaller bound.
Edited copies of one image have different keys, so the service links them in a ledger shared across recipients and records one target per image and assigned forget class.
These copies then reveal no more than after Retrain (Appendix).
The target $\psi^\star(\vx,q)$ in \cref{alg:erase} is this target if there is one and $\psi(q)$ otherwise.

\begin{table*}[t]
\centering
\footnotesize
\renewcommand{\arraystretch}{0.88}
\setlength{\tabcolsep}{1.6pt}
\caption{\textbf{Classification under KR and Normal.} Test accuracy (\%) on \mbox{CIFAR-100} and \mbox{Tiny-ImageNet} with \mbox{ResNet-18}, and ModelNet40 with DGCNN. Bold and underline mark the best and second best unlearning methods, and \ding{51} marks those that read training samples of $\mathcal C_r$.
ERASE and \erasem{} share the gate and the final step, so most of their entries coincide.}
\label{tab:main-classification}
\resizebox{\linewidth}{!}{\begin{tabular}{@{}c@{\hspace{3pt}}lc rrr rrr rrr rrr rrr rrr@{}}
\toprule
 & & & \multicolumn{9}{c}{KR} & \multicolumn{9}{c}{Normal} \\
\cmidrule(lr){4-12}\cmidrule(lr){13-21}
 & & & \multicolumn{3}{c}{$K=1$} & \multicolumn{3}{c}{$K=5$} & \multicolumn{3}{c}{$K=20$} & \multicolumn{3}{c}{$K=1$} & \multicolumn{3}{c}{$K=5$} & \multicolumn{3}{c}{$K=20$} \\
\cmidrule(lr){4-6}\cmidrule(lr){7-9}\cmidrule(lr){10-12}\cmidrule(lr){13-15}\cmidrule(lr){16-18}\cmidrule(lr){19-21}
 & Method & $\mathcal C_r$ & $D_{ft}$$\downarrow$ & $D_{rt}$$\uparrow$ & $\HMt$$\uparrow$ & $D_{ft}$$\downarrow$ & $D_{rt}$$\uparrow$ & $\HMt$$\uparrow$ & $D_{ft}$$\downarrow$ & $D_{rt}$$\uparrow$ & $\HMt$$\uparrow$ & $D_{ft}$$\downarrow$ & $D_{rt}$$\uparrow$ & $\HMt$$\uparrow$ & $D_{ft}$$\downarrow$ & $D_{rt}$$\uparrow$ & $\HMt$$\uparrow$ & $D_{ft}$$\downarrow$ & $D_{rt}$$\uparrow$ & $\HMt$$\uparrow$ \\
\midrule
\multicolumn{21}{@{}l}{\textit{CIFAR-100 / ResNet-18}} \\
 & Original & \multirow{2}{*}{-} & 67.33\pmstd{18.58} & 78.08\pmstd{0.19} & 43.75\pmstd{20.82} & 78.67\pmstd{\hz8.34} & 77.94\pmstd{0.44} & 32.91\pmstd{10.82} & 79.93\pmstd{2.20} & 77.48\pmstd{0.55} & 31.84\pmstd{2.82} & 68.00\pmstd{19.29} & 78.57\pmstd{0.19} & 42.94\pmstd{21.97} & 79.33\pmstd{7.94} & 78.41\pmstd{0.42} & 32.17\pmstd{10.46} & 80.27\pmstd{\hz1.99} & 78.01\pmstd{0.50} & 31.47\pmstd{2.56} \\
 & Retrain & & 80.22\pmstd{13.00} & 76.70\pmstd{0.30} & 29.93\pmstd{16.85} & 78.00\pmstd{\hz6.25} & 73.62\pmstd{0.83} & 33.54\pmstd{\hz7.07} & 68.27\pmstd{0.92} & 68.12\pmstd{0.84} & 43.29\pmstd{0.95} & \hz0.00\pmstd{\hz0.00} & 78.38\pmstd{0.27} & 87.88\pmstd{\hz0.17} & \hz0.00\pmstd{0.00} & 78.50\pmstd{0.55} & 87.95\pmstd{\hz0.35} & \hz0.00\pmstd{\hz0.00} & 79.33\pmstd{0.59} & 88.48\pmstd{0.37} \\
\midrule
\multirow{6}{*}{\rotatebox[origin=c]{90}{\scriptsize Parameter-based}} & Finetune & \multirow{3}{*}{\ding{51}} & 70.67\pmstd{17.29} & 76.04\pmstd{0.58} & 40.19\pmstd{19.89} & 79.60\pmstd{\hz9.13} & 75.83\pmstd{0.39} & 31.44\pmstd{11.48} & 80.03\pmstd{2.60} & 74.66\pmstd{1.11} & 31.46\pmstd{3.36} & \hz8.00\pmstd{\hz5.29} & 75.63\pmstd{0.48} & 82.98\pmstd{\hz2.44} & \hz6.13\pmstd{3.40} & 75.68\pmstd{0.97} & 83.79\pmstd{\hz1.90} & 13.20\pmstd{\hz2.55} & 77.99\pmstd{1.06} & 82.15\pmstd{1.53} \\
 & SalUn & & 84.89\pmstd{11.97} & 76.86\pmstd{0.32} & 23.89\pmstd{15.98} & 78.27\pmstd{\hz9.40} & 73.88\pmstd{0.46} & 32.81\pmstd{11.54} & 70.59\pmstd{1.66} & 71.23\pmstd{0.44} & 41.62\pmstd{1.75} & 10.00\pmstd{\hz6.93} & 77.52\pmstd{0.48} & 83.22\pmstd{\hz3.18} & \hz1.73\pmstd{0.46} & 75.58\pmstd{0.04} & 85.44\pmstd{\hz0.19} & \hz\underline{0.77\pmstd{\hz0.15}} & 75.28\pmstd{1.58} & 85.61\pmstd{1.06} \\
 & BadT & & 73.11\pmstd{18.81} & 77.45\pmstd{0.05} & 37.08\pmstd{23.57} & 58.18\pmstd{\hz5.30} & 76.57\pmstd{0.16} & 53.96\pmstd{\hz4.29} & 39.80\pmstd{3.16} & 72.91\pmstd{0.85} & 65.91\pmstd{1.80} & \hz\textbf{0.00\pmstd{\hz0.00}} & 77.53\pmstd{0.41} & 87.34\pmstd{\hz0.26} & \hz\textbf{0.00\pmstd{0.00}} & 76.40\pmstd{0.36} & 86.62\pmstd{\hz0.23} & \hz\textbf{0.00\pmstd{\hz0.00}} & 69.33\pmstd{0.34} & 81.89\pmstd{0.24} \\
\cmidrule(l){2-21}
 & NG & \multirow{3}{*}{\ding{55}} & 65.56\pmstd{16.06} & 75.48\pmstd{1.07} & 45.73\pmstd{16.85} & 62.89\pmstd{10.82} & 72.29\pmstd{2.04} & 48.34\pmstd{\hz9.97} & 73.52\pmstd{4.22} & 73.83\pmstd{2.60} & 38.80\pmstd{4.13} & \hz\underline{1.33\pmstd{\hz1.15}} & 71.87\pmstd{0.85} & 83.16\pmstd{\hz0.93} & \hz\textbf{0.00\pmstd{0.00}} & 66.03\pmstd{4.30} & 79.49\pmstd{\hz3.08} & 31.07\pmstd{12.26} & 71.43\pmstd{5.91} & 69.38\pmstd{3.64} \\
 & RL & & 65.33\pmstd{16.22} & 75.51\pmstd{1.07} & 45.92\pmstd{16.98} & 60.84\pmstd{14.34} & 67.75\pmstd{1.12} & 48.50\pmstd{12.51} & 39.48\pmstd{5.18} & 57.54\pmstd{3.40} & 58.81\pmstd{1.39} & \hz\underline{1.33\pmstd{\hz1.15}} & 71.84\pmstd{0.91} & 83.14\pmstd{\hz0.95} & \hz3.33\pmstd{0.23} & 64.95\pmstd{1.67} & 77.69\pmstd{\hz1.23} & \hz3.80\pmstd{\hz0.87} & 52.33\pmstd{4.23} & 67.71\pmstd{3.42} \\
 & DELETE & & 67.56\pmstd{16.55} & 76.29\pmstd{0.91} & 43.79\pmstd{17.87} & 77.78\pmstd{\hz8.73} & 76.86\pmstd{0.48} & 33.86\pmstd{10.84} & 75.06\pmstd{1.28} & 73.93\pmstd{0.30} & 37.29\pmstd{1.46} & \hz\underline{1.33\pmstd{\hz1.15}} & 73.91\pmstd{1.20} & 84.51\pmstd{\hz1.05} & \hz1.87\pmstd{1.85} & 77.72\pmstd{1.07} & 86.74\pmstd{\hz1.37} & \hz4.87\pmstd{\hz1.99} & 78.10\pmstd{0.54} & 85.77\pmstd{0.80} \\
\midrule
\multirow{6}{*}{\rotatebox[origin=c]{90}{\scriptsize Feature-level}} & R-ESC & \multirow{2}{*}{\ding{51}} & \hz\underline{7.33\pmstd{\hz4.67}} & \underline{77.90\pmstd{0.15}} & \underline{84.61\pmstd{\hz2.05}} & \underline{10.98\pmstd{\hz2.21}} & 75.91\pmstd{0.25} & \underline{81.94\pmstd{\hz0.90}} & \hz1.11\pmstd{0.31} & 69.58\pmstd{0.38} & 81.68\pmstd{0.17} & \hz\textbf{0.00\pmstd{\hz0.00}} & 78.46\pmstd{0.15} & 87.93\pmstd{\hz0.09} & \hz\textbf{0.00\pmstd{0.00}} & 78.37\pmstd{0.37} & 87.87\pmstd{\hz0.23} & \hz\textbf{0.00\pmstd{\hz0.00}} & 77.02\pmstd{0.48} & 87.02\pmstd{0.31} \\
 & LOFT & & 67.33\pmstd{23.01} & 76.16\pmstd{0.32} & 42.43\pmstd{23.52} & 34.89\pmstd{\hz7.46} & 76.58\pmstd{0.31} & 70.23\pmstd{\hz4.59} & 56.32\pmstd{0.67} & 71.47\pmstd{0.88} & 54.21\pmstd{0.42} & \hz\textbf{0.00\pmstd{\hz0.00}} & 68.26\pmstd{1.63} & 81.13\pmstd{\hz1.16} & \hz\textbf{0.00\pmstd{0.00}} & 77.45\pmstd{0.28} & 87.29\pmstd{\hz0.18} & \hz\textbf{0.00\pmstd{\hz0.00}} & \underline{78.58\pmstd{0.87}} & \underline{88.01\pmstd{0.55}} \\
\cmidrule(l){2-21}
 & ESC & \multirow{2}{*}{\ding{55}} & 59.11\pmstd{13.00} & 77.89\pmstd{0.08} & 52.76\pmstd{12.00} & 14.71\pmstd{\hz5.95} & \underline{77.86\pmstd{0.31}} & 81.35\pmstd{\hz2.92} & 12.82\pmstd{1.18} & 76.73\pmstd{0.78} & 81.62\pmstd{0.93} & \hz2.00\pmstd{\hz2.00} & 78.01\pmstd{0.27} & 86.86\pmstd{\hz0.69} & \hz\underline{0.13\pmstd{0.23}} & 78.02\pmstd{0.51} & 87.60\pmstd{\hz0.41} & \hz\textbf{0.00\pmstd{\hz0.00}} & 77.73\pmstd{1.33} & 87.46\pmstd{0.84} \\
 & POUR-P & & 75.11\pmstd{\hz8.95} & 77.62\pmstd{0.16} & 37.07\pmstd{10.83} & 79.60\pmstd{\hz7.08} & 76.64\pmstd{0.22} & 31.78\pmstd{\hz9.20} & 77.89\pmstd{0.73} & 74.65\pmstd{0.58} & 34.11\pmstd{0.93} & \hz\textbf{0.00\pmstd{\hz0.00}} & \underline{78.52\pmstd{0.18}} & \underline{87.97\pmstd{\hz0.11}} & \hz\textbf{0.00\pmstd{0.00}} & \textbf{78.76\pmstd{0.46}} & \textbf{88.12\pmstd{\hz0.29}} & \hz\textbf{0.00\pmstd{\hz0.00}} & \textbf{79.05\pmstd{0.78}} & \textbf{88.30\pmstd{0.49}} \\
 & \cellcolor{eraserow}\textbf{ERASE} & \cellcolor{eraserow}\ding{55} & \cellcolor{eraserow}\hz\textbf{0.00\pmstd{\hz0.00}} & \cellcolor{eraserow}\textbf{78.57\pmstd{0.20}} & \cellcolor{eraserow}\textbf{88.00\pmstd{\hz0.13}} & \cellcolor{eraserow}\hz\textbf{0.00\pmstd{\hz0.00}} & \cellcolor{eraserow}\textbf{78.42\pmstd{0.41}} & \cellcolor{eraserow}\textbf{87.91\pmstd{\hz0.26}} & \cellcolor{eraserow}\hz\underline{0.06\pmstd{0.04}} & \cellcolor{eraserow}\underline{78.07\pmstd{0.51}} & \cellcolor{eraserow}\underline{87.66\pmstd{0.33}} & \cellcolor{eraserow}\hz\textbf{0.00\pmstd{\hz0.00}} & \cellcolor{eraserow}\textbf{78.57\pmstd{0.20}} & \cellcolor{eraserow}\textbf{88.00\pmstd{\hz0.13}} & \cellcolor{eraserow}\hz\textbf{0.00\pmstd{0.00}} & \cellcolor{eraserow}\underline{78.42\pmstd{0.41}} & \cellcolor{eraserow}\underline{87.91\pmstd{\hz0.26}} & \cellcolor{eraserow}\hz\textbf{0.00\pmstd{\hz0.00}} & \cellcolor{eraserow}78.08\pmstd{0.52} & \cellcolor{eraserow}87.69\pmstd{0.33} \\
 & \cellcolor{erasemrow}\textbf{ERASE-M} & \cellcolor{erasemrow}\ding{55} & \cellcolor{erasemrow}\hz\textbf{0.00\pmstd{\hz0.00}} & \cellcolor{erasemrow}\textbf{78.57\pmstd{0.20}} & \cellcolor{erasemrow}\textbf{88.00\pmstd{\hz0.13}} & \cellcolor{erasemrow}\hz\textbf{0.00\pmstd{\hz0.00}} & \cellcolor{erasemrow}\textbf{78.42\pmstd{0.41}} & \cellcolor{erasemrow}\textbf{87.91\pmstd{\hz0.26}} & \cellcolor{erasemrow}\hz\textbf{0.00\pmstd{0.00}} & \cellcolor{erasemrow}\textbf{78.08\pmstd{0.52}} & \cellcolor{erasemrow}\textbf{87.69\pmstd{0.33}} & \cellcolor{erasemrow}\hz\textbf{0.00\pmstd{\hz0.00}} & \cellcolor{erasemrow}\textbf{78.57\pmstd{0.20}} & \cellcolor{erasemrow}\textbf{88.00\pmstd{\hz0.13}} & \cellcolor{erasemrow}\hz\textbf{0.00\pmstd{0.00}} & \cellcolor{erasemrow}\underline{78.42\pmstd{0.41}} & \cellcolor{erasemrow}\underline{87.91\pmstd{\hz0.26}} & \cellcolor{erasemrow}\hz\textbf{0.00\pmstd{\hz0.00}} & \cellcolor{erasemrow}78.08\pmstd{0.52} & \cellcolor{erasemrow}87.69\pmstd{0.33} \\
\midrule
\multicolumn{21}{@{}l}{\textit{Tiny-ImageNet / ResNet-18}} \\
 & Original & \multirow{2}{*}{-} & 65.33\pmstd{10.58} & 64.36\pmstd{0.05} & 44.45\pmstd{8.59} & 61.87\pmstd{\hz8.12} & 64.42\pmstd{0.21} & 47.57\pmstd{6.65} & 66.44\pmstd{2.63} & 64.13\pmstd{0.29} & 44.02\pmstd{2.34} & 66.67\pmstd{10.07} & 64.39\pmstd{0.05} & 43.34\pmstd{8.68} & 63.73\pmstd{8.99} & 64.42\pmstd{0.23} & 45.96\pmstd{7.70} & 66.93\pmstd{\hz3.40} & 64.12\pmstd{0.38} & 43.57\pmstd{3.06} \\
 & Retrain & & 72.89\pmstd{\hz7.58} & 62.99\pmstd{0.47} & 37.48\pmstd{7.72} & 67.11\pmstd{\hz5.60} & 60.37\pmstd{0.57} & 42.38\pmstd{4.45} & 63.56\pmstd{4.36} & 56.61\pmstd{0.24} & 44.24\pmstd{3.08} & \hz0.00\pmstd{\hz0.00} & 64.08\pmstd{0.33} & 78.11\pmstd{0.25} & \hz0.00\pmstd{0.00} & 64.22\pmstd{0.10} & 78.21\pmstd{0.07} & \hz0.00\pmstd{\hz0.00} & 64.99\pmstd{0.21} & 78.78\pmstd{0.15} \\
\midrule
\multirow{6}{*}{\rotatebox[origin=c]{90}{\scriptsize Parameter-based}} & Finetune & \multirow{3}{*}{\ding{51}} & 71.11\pmstd{\hz3.36} & 61.68\pmstd{0.35} & 39.28\pmstd{3.20} & 66.58\pmstd{\hz8.87} & 61.68\pmstd{0.40} & 42.90\pmstd{7.15} & 67.76\pmstd{3.61} & 60.28\pmstd{0.20} & 41.94\pmstd{3.05} & \hz\textbf{0.00\pmstd{\hz0.00}} & 61.53\pmstd{0.14} & 76.18\pmstd{0.11} & \hz2.67\pmstd{2.57} & 62.26\pmstd{0.62} & 75.94\pmstd{1.21} & \hz2.47\pmstd{\hz0.92} & 62.90\pmstd{0.50} & 76.48\pmstd{0.65} \\
 & SalUn & & 68.00\pmstd{\hz4.62} & 63.84\pmstd{0.08} & 42.50\pmstd{4.00} & 64.89\pmstd{\hz8.47} & 60.40\pmstd{0.46} & 43.96\pmstd{7.09} & 60.36\pmstd{3.78} & 59.76\pmstd{0.18} & 47.60\pmstd{2.74} & \hz5.33\pmstd{\hz2.31} & 63.95\pmstd{0.23} & 76.32\pmstd{0.65} & \hz2.67\pmstd{1.67} & 61.72\pmstd{0.37} & 75.53\pmstd{0.63} & \hz7.80\pmstd{\hz1.78} & 62.70\pmstd{0.17} & 74.63\pmstd{0.69} \\
 & BadT & & 64.44\pmstd{\hz3.85} & 63.93\pmstd{0.05} & 45.62\pmstd{3.10} & 48.18\pmstd{\hz1.26} & 60.85\pmstd{0.12} & 55.97\pmstd{0.69} & 38.60\pmstd{1.76} & 59.58\pmstd{0.98} & 60.46\pmstd{0.58} & \hz5.33\pmstd{\hz2.31} & 63.79\pmstd{0.28} & 76.21\pmstd{0.65} & \hz\textbf{0.00\pmstd{0.00}} & 60.60\pmstd{0.28} & 75.47\pmstd{0.22} & \hz\underline{0.80\pmstd{\hz1.39}} & 59.36\pmstd{0.94} & 74.27\pmstd{0.34} \\
\cmidrule(l){2-21}
 & NG & \multirow{3}{*}{\ding{55}} & 64.89\pmstd{11.18} & \underline{64.15\pmstd{0.33}} & 44.73\pmstd{8.89} & 44.62\pmstd{12.27} & 59.16\pmstd{1.26} & 56.64\pmstd{6.67} & 58.98\pmstd{6.15} & 60.95\pmstd{2.13} & 48.77\pmstd{4.03} & 52.00\pmstd{\hz8.00} & 64.15\pmstd{0.56} & 54.64\pmstd{5.06} & \hz\underline{0.27\pmstd{0.46}} & 54.94\pmstd{1.41} & 70.84\pmstd{1.19} & 27.73\pmstd{11.65} & 58.89\pmstd{3.53} & 64.36\pmstd{3.14} \\
 & RL & & 59.56\pmstd{\hz3.36} & 59.11\pmstd{1.53} & 47.94\pmstd{1.84} & 52.62\pmstd{\hz7.79} & 60.89\pmstd{0.54} & 53.03\pmstd{5.05} & 49.51\pmstd{2.98} & 55.88\pmstd{0.57} & 53.01\pmstd{1.61} & \hz\underline{1.33\pmstd{\hz2.31}} & 56.25\pmstd{1.91} & 71.65\pmstd{2.13} & \hz4.27\pmstd{1.85} & 59.21\pmstd{0.47} & 73.16\pmstd{0.53} & \hz6.13\pmstd{\hz0.83} & 53.34\pmstd{1.34} & 68.01\pmstd{0.88} \\
 & DELETE & & 63.56\pmstd{\hz7.81} & 61.85\pmstd{0.72} & 45.51\pmstd{5.76} & 56.80\pmstd{10.54} & 63.79\pmstd{0.25} & 51.01\pmstd{8.02} & 63.36\pmstd{2.94} & 62.98\pmstd{0.38} & 46.29\pmstd{2.48} & \hz\textbf{0.00\pmstd{\hz0.00}} & 58.26\pmstd{1.73} & 73.61\pmstd{1.39} & \hz\textbf{0.00\pmstd{0.00}} & 64.06\pmstd{0.43} & 78.09\pmstd{0.32} & \hz2.60\pmstd{\hz0.72} & \textbf{64.56\pmstd{0.62}} & 77.65\pmstd{0.36} \\
\midrule
\multirow{6}{*}{\rotatebox[origin=c]{90}{\scriptsize Feature-level}} & R-ESC & \multirow{2}{*}{\ding{51}} & \underline{12.00\pmstd{\hz7.42}} & 64.04\pmstd{0.12} & \underline{74.05\pmstd{2.70}} & \hz\underline{4.44\pmstd{\hz0.56}} & 59.23\pmstd{0.40} & 73.13\pmstd{0.14} & \hz1.11\pmstd{0.25} & 56.49\pmstd{0.26} & 71.90\pmstd{0.26} & \hz\textbf{0.00\pmstd{\hz0.00}} & 64.34\pmstd{0.09} & 78.30\pmstd{0.06} & \hz\textbf{0.00\pmstd{0.00}} & 63.95\pmstd{0.27} & 78.01\pmstd{0.20} & \hz\textbf{0.00\pmstd{\hz0.00}} & 61.61\pmstd{0.31} & 76.24\pmstd{0.24} \\
 & LOFT & & 60.44\pmstd{\hz7.70} & 61.31\pmstd{0.33} & 47.79\pmstd{5.99} & 18.22\pmstd{\hz2.48} & 62.85\pmstd{0.40} & 71.06\pmstd{0.67} & 27.98\pmstd{2.48} & 59.37\pmstd{0.43} & 65.07\pmstd{0.96} & \hz\textbf{0.00\pmstd{\hz0.00}} & 52.15\pmstd{0.82} & 68.55\pmstd{0.71} & \hz\textbf{0.00\pmstd{0.00}} & 59.90\pmstd{0.16} & 74.92\pmstd{0.13} & \hz\textbf{0.00\pmstd{\hz0.00}} & 60.65\pmstd{0.54} & 75.51\pmstd{0.42} \\
\cmidrule(l){2-21}
 & ESC & \multirow{2}{*}{\ding{55}} & 55.11\pmstd{\hz4.29} & \underline{64.15\pmstd{0.07}} & 52.74\pmstd{3.05} & \hz9.24\pmstd{\hz1.20} & \underline{63.98\pmstd{0.08}} & \underline{75.05\pmstd{0.46}} & 11.73\pmstd{1.46} & 61.40\pmstd{0.13} & 72.42\pmstd{0.54} & \hz4.00\pmstd{\hz4.00} & 62.83\pmstd{0.23} & 75.93\pmstd{1.42} & \hz\textbf{0.00\pmstd{0.00}} & 63.17\pmstd{0.21} & 77.43\pmstd{0.16} & \hz\textbf{0.00\pmstd{\hz0.00}} & 60.96\pmstd{0.57} & 75.74\pmstd{0.44} \\
 & POUR-P & & 72.89\pmstd{\hz5.05} & 63.98\pmstd{0.07} & 37.89\pmstd{5.14} & 62.13\pmstd{\hz5.54} & 63.15\pmstd{0.12} & 47.18\pmstd{4.50} & 64.20\pmstd{2.36} & 61.05\pmstd{0.17} & 45.11\pmstd{1.91} & \hz\textbf{0.00\pmstd{\hz0.00}} & \underline{64.38\pmstd{0.15}} & \underline{78.33\pmstd{0.11}} & \hz\textbf{0.00\pmstd{0.00}} & \underline{64.36\pmstd{0.37}} & \underline{78.31\pmstd{0.27}} & \hz\textbf{0.00\pmstd{\hz0.00}} & 63.87\pmstd{0.64} & \underline{77.95\pmstd{0.48}} \\
 & \cellcolor{eraserow}\textbf{ERASE} & \cellcolor{eraserow}\ding{55} & \cellcolor{eraserow}\hz\textbf{0.00\pmstd{\hz0.00}} & \cellcolor{eraserow}\textbf{64.40\pmstd{0.04}} & \cellcolor{eraserow}\textbf{78.34\pmstd{0.03}} & \cellcolor{eraserow}\hz\textbf{0.00\pmstd{\hz0.00}} & \cellcolor{eraserow}\textbf{64.42\pmstd{0.23}} & \cellcolor{eraserow}\textbf{78.36\pmstd{0.17}} & \cellcolor{eraserow}\hz\underline{0.09\pmstd{0.10}} & \cellcolor{eraserow}\underline{64.12\pmstd{0.37}} & \cellcolor{eraserow}\underline{78.11\pmstd{0.29}} & \cellcolor{eraserow}\hz\textbf{0.00\pmstd{\hz0.00}} & \cellcolor{eraserow}\textbf{64.40\pmstd{0.04}} & \cellcolor{eraserow}\textbf{78.34\pmstd{0.03}} & \cellcolor{eraserow}\hz\textbf{0.00\pmstd{0.00}} & \cellcolor{eraserow}\textbf{64.42\pmstd{0.23}} & \cellcolor{eraserow}\textbf{78.36\pmstd{0.17}} & \cellcolor{eraserow}\hz\textbf{0.00\pmstd{\hz0.00}} & \cellcolor{eraserow}\underline{64.13\pmstd{0.38}} & \cellcolor{eraserow}\textbf{78.14\pmstd{0.28}} \\
 & \cellcolor{erasemrow}\textbf{ERASE-M} & \cellcolor{erasemrow}\ding{55} & \cellcolor{erasemrow}\hz\textbf{0.00\pmstd{\hz0.00}} & \cellcolor{erasemrow}\textbf{64.40\pmstd{0.04}} & \cellcolor{erasemrow}\textbf{78.34\pmstd{0.03}} & \cellcolor{erasemrow}\hz\textbf{0.00\pmstd{\hz0.00}} & \cellcolor{erasemrow}\textbf{64.42\pmstd{0.23}} & \cellcolor{erasemrow}\textbf{78.36\pmstd{0.17}} & \cellcolor{erasemrow}\hz\textbf{0.00\pmstd{0.00}} & \cellcolor{erasemrow}\textbf{64.13\pmstd{0.38}} & \cellcolor{erasemrow}\textbf{78.14\pmstd{0.28}} & \cellcolor{erasemrow}\hz\textbf{0.00\pmstd{\hz0.00}} & \cellcolor{erasemrow}\textbf{64.40\pmstd{0.04}} & \cellcolor{erasemrow}\textbf{78.34\pmstd{0.03}} & \cellcolor{erasemrow}\hz\textbf{0.00\pmstd{0.00}} & \cellcolor{erasemrow}\textbf{64.42\pmstd{0.23}} & \cellcolor{erasemrow}\textbf{78.36\pmstd{0.17}} & \cellcolor{erasemrow}\hz\textbf{0.00\pmstd{\hz0.00}} & \cellcolor{erasemrow}\underline{64.13\pmstd{0.38}} & \cellcolor{erasemrow}\textbf{78.14\pmstd{0.28}} \\
\midrule
\multicolumn{21}{@{}l}{\textit{ModelNet40 / DGCNN}} \\
 & & & \multicolumn{3}{c}{$K=1$} & \multicolumn{3}{c}{$K=5$} & \multicolumn{3}{c}{$K=10$} & \multicolumn{3}{c}{$K=1$} & \multicolumn{3}{c}{$K=5$} & \multicolumn{3}{c}{$K=10$} \\
\cmidrule(lr){4-6}\cmidrule(lr){7-9}\cmidrule(lr){10-12}\cmidrule(lr){13-15}\cmidrule(lr){16-18}\cmidrule(lr){19-21}
 & Original & \multirow{2}{*}{-} & 93.78\pmstd{\hz4.02} & 92.55\pmstd{0.13} & 11.47\pmstd{\hz7.12} & 92.75\pmstd{4.75} & 92.61\pmstd{0.71} & 13.21\pmstd{8.00} & 92.47\pmstd{1.95} & 92.69\pmstd{0.85} & 13.89\pmstd{3.32} & 98.00\pmstd{\hz3.46} & 93.07\pmstd{0.14} & \hz3.76\pmstd{\hz6.51} & 93.34\pmstd{\hz4.46} & 93.19\pmstd{0.65} & 12.20\pmstd{\hz7.65} & 92.96\pmstd{\hz1.96} & 93.32\pmstd{0.80} & 13.05\pmstd{\hz3.36} \\
 & Retrain & & 76.89\pmstd{17.52} & 92.33\pmstd{0.76} & 34.89\pmstd{20.99} & 80.02\pmstd{2.59} & 90.40\pmstd{0.88} & 32.67\pmstd{3.50} & 76.39\pmstd{3.18} & 90.43\pmstd{0.80} & 37.37\pmstd{4.07} & \hz0.00\pmstd{\hz0.00} & 93.51\pmstd{0.34} & 96.65\pmstd{\hz0.18} & \hz0.00\pmstd{\hz0.00} & 93.84\pmstd{0.52} & 96.82\pmstd{\hz0.28} & \hz0.00\pmstd{\hz0.00} & 95.63\pmstd{1.26} & 97.77\pmstd{\hz0.66} \\
\midrule
\multirow{6}{*}{\rotatebox[origin=c]{90}{\scriptsize Parameter-based}} & Finetune & \multirow{3}{*}{\ding{51}} & 91.33\pmstd{\hz4.16} & \underline{92.81\pmstd{0.20}} & 15.66\pmstd{\hz7.10} & 90.37\pmstd{3.19} & 92.74\pmstd{0.72} & 17.34\pmstd{5.24} & 89.94\pmstd{1.40} & 93.23\pmstd{0.48} & 18.14\pmstd{2.28} & 60.00\pmstd{38.16} & 92.60\pmstd{0.71} & 49.53\pmstd{33.83} & 63.27\pmstd{11.50} & 93.55\pmstd{0.67} & 52.09\pmstd{11.36} & 62.65\pmstd{10.38} & 94.47\pmstd{1.10} & 52.99\pmstd{10.28} \\
 & SalUn & & 91.56\pmstd{\hz4.54} & 92.60\pmstd{0.35} & 15.25\pmstd{\hz7.80} & 92.44\pmstd{4.45} & 92.77\pmstd{0.74} & 13.77\pmstd{7.51} & 90.92\pmstd{1.02} & 92.89\pmstd{0.21} & 16.53\pmstd{1.70} & 50.00\pmstd{44.00} & 92.21\pmstd{0.33} & 56.44\pmstd{41.64} & 76.53\pmstd{12.49} & 93.20\pmstd{0.78} & 36.36\pmstd{16.44} & 78.04\pmstd{\hz5.94} & 94.48\pmstd{0.29} & 35.37\pmstd{\hz7.80} \\
 & BadT & & 93.11\pmstd{\hz3.79} & 92.49\pmstd{0.46} & 12.65\pmstd{\hz6.68} & 87.88\pmstd{2.96} & 92.81\pmstd{0.69} & 21.37\pmstd{4.58} & 86.59\pmstd{3.04} & 93.05\pmstd{0.35} & 23.36\pmstd{4.65} & 56.00\pmstd{10.58} & \underline{93.12\pmstd{0.43}} & 59.30\pmstd{\hz9.55} & 58.08\pmstd{\hz8.81} & \underline{93.68\pmstd{0.55}} & 57.56\pmstd{\hz8.79} & 60.95\pmstd{\hz3.12} & 94.43\pmstd{0.81} & 55.21\pmstd{\hz3.27} \\
\cmidrule(l){2-21}
 & NG & \multirow{3}{*}{\ding{55}} & 84.89\pmstd{\hz1.39} & 92.32\pmstd{0.33} & 25.95\pmstd{\hz2.04} & 89.74\pmstd{2.61} & 92.17\pmstd{0.84} & 18.39\pmstd{4.22} & 89.36\pmstd{0.41} & 92.79\pmstd{0.77} & 19.08\pmstd{0.68} & 17.33\pmstd{\hz4.62} & 91.80\pmstd{0.68} & 86.95\pmstd{\hz2.67} & 34.62\pmstd{12.01} & 90.20\pmstd{1.71} & 75.35\pmstd{\hz8.11} & 47.17\pmstd{\hz5.08} & 91.23\pmstd{1.87} & 66.82\pmstd{\hz4.29} \\
 & RL & & 82.89\pmstd{\hz3.42} & 92.25\pmstd{0.30} & 28.77\pmstd{\hz4.93} & 91.25\pmstd{3.27} & 92.60\pmstd{0.96} & 15.88\pmstd{5.39} & 92.02\pmstd{1.72} & 92.93\pmstd{0.83} & 14.66\pmstd{2.90} & 37.33\pmstd{\hz7.02} & 91.30\pmstd{0.85} & 74.14\pmstd{\hz4.63} & 81.39\pmstd{\hz6.22} & 92.74\pmstd{0.90} & 30.68\pmstd{\hz8.96} & 88.82\pmstd{\hz2.47} & 93.66\pmstd{0.80} & 19.91\pmstd{\hz3.97} \\
 & DELETE & & 74.00\pmstd{20.95} & 92.16\pmstd{0.16} & 37.75\pmstd{23.80} & 86.11\pmstd{3.89} & 91.91\pmstd{0.83} & 23.99\pmstd{5.84} & 78.59\pmstd{4.18} & 91.61\pmstd{1.07} & 34.59\pmstd{5.55} & 10.00\pmstd{\hz2.00} & 90.09\pmstd{0.91} & 90.04\pmstd{\hz1.43} & \hz\underline{8.53\pmstd{\hz3.87}} & 92.71\pmstd{0.36} & 92.06\pmstd{\hz2.09} & 15.96\pmstd{\hz7.25} & 93.45\pmstd{0.77} & 88.41\pmstd{\hz4.39} \\
\midrule
\multirow{6}{*}{\rotatebox[origin=c]{90}{\scriptsize Feature-level}} & R-ESC & \multirow{2}{*}{\ding{51}} & \hz4.22\pmstd{\hz6.74} & 91.72\pmstd{0.76} & 93.64\pmstd{\hz3.56} & \hz0.62\pmstd{0.41} & 89.43\pmstd{0.47} & 94.14\pmstd{0.35} & \hz0.67\pmstd{0.79} & 89.07\pmstd{0.39} & 93.92\pmstd{0.52} & \hz\textbf{0.00\pmstd{\hz0.00}} & 93.04\pmstd{0.12} & \underline{96.39\pmstd{\hz0.06}} & \hz\textbf{0.00\pmstd{\hz0.00}} & 92.81\pmstd{0.57} & 96.27\pmstd{\hz0.30} & \hz\textbf{0.00\pmstd{\hz0.00}} & 93.25\pmstd{1.17} & 96.50\pmstd{\hz0.62} \\
 & LOFT & & 72.44\pmstd{42.54} & 92.20\pmstd{0.42} & 31.83\pmstd{45.14} & 33.31\pmstd{8.75} & 91.53\pmstd{0.66} & 76.94\pmstd{6.04} & 16.51\pmstd{4.47} & 92.60\pmstd{0.63} & 87.77\pmstd{2.63} & \hz4.67\pmstd{\hz8.08} & 89.72\pmstd{1.13} & 92.36\pmstd{\hz4.47} & 23.37\pmstd{\hz6.95} & 93.27\pmstd{0.50} & 84.01\pmstd{\hz3.99} & 10.43\pmstd{\hz4.47} & \underline{94.55\pmstd{1.07}} & 91.94\pmstd{\hz1.93} \\
\cmidrule(l){2-21}
 & ESC & \multirow{2}{*}{\ding{55}} & 56.67\pmstd{49.10} & 92.63\pmstd{0.28} & 49.24\pmstd{40.70} & 31.10\pmstd{6.40} & \underline{92.99\pmstd{0.49}} & 79.04\pmstd{4.23} & 15.66\pmstd{4.58} & 93.18\pmstd{0.22} & 88.50\pmstd{2.60} & \hz\underline{0.67\pmstd{\hz1.15}} & 92.82\pmstd{0.45} & 95.96\pmstd{\hz0.31} & \hz\textbf{0.00\pmstd{\hz0.00}} & 92.96\pmstd{0.80} & \underline{96.35\pmstd{\hz0.43}} & \hz\textbf{0.00\pmstd{\hz0.00}} & 93.55\pmstd{0.60} & \textbf{96.67\pmstd{\hz0.32}} \\
 & POUR-P & & 87.78\pmstd{10.69} & 92.43\pmstd{0.35} & 20.46\pmstd{16.68} & 90.93\pmstd{3.92} & 92.37\pmstd{0.74} & 16.37\pmstd{6.41} & 90.08\pmstd{2.45} & 92.58\pmstd{1.01} & 17.85\pmstd{4.06} & \hz6.67\pmstd{11.55} & \textbf{93.21\pmstd{0.30}} & 93.01\pmstd{\hz5.88} & 18.52\pmstd{10.83} & \textbf{93.75\pmstd{0.56}} & 86.90\pmstd{\hz6.11} & \hz\underline{7.96\pmstd{\hz5.48}} & \textbf{94.58\pmstd{0.98}} & 93.22\pmstd{\hz2.34} \\
 & \cellcolor{eraserow}\textbf{ERASE} & \cellcolor{eraserow}\ding{55} & \cellcolor{eraserow}\hz\textbf{0.00\pmstd{\hz0.00}} & \cellcolor{eraserow}\textbf{93.07\pmstd{0.14}} & \cellcolor{eraserow}\textbf{96.41\pmstd{\hz0.08}} & \cellcolor{eraserow}\hz\underline{0.05\pmstd{0.09}} & \cellcolor{eraserow}\textbf{93.19\pmstd{0.65}} & \cellcolor{eraserow}\underline{96.45\pmstd{0.37}} & \cellcolor{eraserow}\hz\underline{0.09\pmstd{0.15}} & \cellcolor{eraserow}\underline{93.31\pmstd{0.77}} & \cellcolor{eraserow}\underline{96.50\pmstd{0.45}} & \cellcolor{eraserow}\hz\textbf{0.00\pmstd{\hz0.00}} & \cellcolor{eraserow}93.07\pmstd{0.14} & \cellcolor{eraserow}\textbf{96.41\pmstd{\hz0.08}} & \cellcolor{eraserow}\hz\textbf{0.00\pmstd{\hz0.00}} & \cellcolor{eraserow}93.19\pmstd{0.65} & \cellcolor{eraserow}\textbf{96.47\pmstd{\hz0.35}} & \cellcolor{eraserow}\hz\textbf{0.00\pmstd{\hz0.00}} & \cellcolor{eraserow}93.36\pmstd{0.76} & \cellcolor{eraserow}\underline{96.57\pmstd{\hz0.40}} \\
 & \cellcolor{erasemrow}\textbf{ERASE-M} & \cellcolor{erasemrow}\ding{55} & \cellcolor{erasemrow}\hz\underline{0.67\pmstd{\hz1.15}} & \cellcolor{erasemrow}\textbf{93.07\pmstd{0.14}} & \cellcolor{erasemrow}\underline{96.10\pmstd{\hz0.50}} & \cellcolor{erasemrow}\hz\textbf{0.00\pmstd{0.00}} & \cellcolor{erasemrow}\textbf{93.19\pmstd{0.65}} & \cellcolor{erasemrow}\textbf{96.47\pmstd{0.35}} & \cellcolor{erasemrow}\hz\textbf{0.00\pmstd{0.00}} & \cellcolor{erasemrow}\textbf{93.36\pmstd{0.86}} & \cellcolor{erasemrow}\textbf{96.57\pmstd{0.46}} & \cellcolor{erasemrow}\hz\textbf{0.00\pmstd{\hz0.00}} & \cellcolor{erasemrow}93.07\pmstd{0.14} & \cellcolor{erasemrow}\textbf{96.41\pmstd{\hz0.08}} & \cellcolor{erasemrow}\hz\textbf{0.00\pmstd{\hz0.00}} & \cellcolor{erasemrow}93.19\pmstd{0.65} & \cellcolor{erasemrow}\textbf{96.47\pmstd{\hz0.35}} & \cellcolor{erasemrow}\hz\textbf{0.00\pmstd{\hz0.00}} & \cellcolor{erasemrow}93.36\pmstd{0.86} & \cellcolor{erasemrow}\underline{96.57\pmstd{\hz0.46}} \\
\bottomrule
\end{tabular}}
\end{table*}

\subsection{Predicted Target Distribution (ERASE)}
\label{sec:replacement}
Once the gate assigns a target $k$, ERASE needs the location and shape of its distribution.
Without remaining-class samples, it predicts both relative to the center of the remaining classes.
Before a deletion request, the service stores the class-balanced mean $\bar\vz=\frac{1}{C}\sum\nolimits_{j\in\cY}\vmu_j$ of the training features and the class-balanced covariance $\boldsymbol\Sigma$ of their projections about $\mP\bar\vz$.
Since the projection is linear, removing the forget-class means gives this center, the mean of the projected remaining-class means, \mbox{$\hat\vmu_r=\frac{1}{C-K}\bigl(C\mP\bar\vz-\sum\nolimits_{c\in\cCf}\mP\vmu_c\bigr)$}.

ERASE places each remaining class at this center plus a step along a direction set by its head row, with a step length fitted to the known forget-class means.
Two cases tie the rows to the centered class means.
In collapsed networks the means align with the rows~\cite{papyan2020prevalence}, and for a class-balanced least-squares head with an intercept~\cite{hastie2009elements} they satisfy $C\boldsymbol\Sigma\vw_j=\mP\vmu_j-\mP\bar\vz$.
ERASE spans both cases with a power $\alpha\in[0,1]$ of $\boldsymbol\Sigma$ applied to templates.
The template $\vt_j$ is the smallest nonnegative multiple of $\vw_j$ that the head assigns to class $j$ with probability at least $1-\epsilon$, for $\epsilon\in(0,\frac12)$.
The predicted location of a remaining class $k$ is
\begin{equation*}
\boldsymbol\theta_k=\hat\vmu_r+\lambda\Bigl(\boldsymbol\Sigma^{\alpha}\vt_k-\frac{1}{C-K}\sum_{l\in\mathcal C_r}\boldsymbol\Sigma^{\alpha}\vt_l\Bigr),
\end{equation*}
where one signed coefficient $\lambda$ fits the known forget-class means by least squares and the stored moments choose $\alpha$.

For the shape, ERASE borrows the shape of the forget classes.
Let $\boldsymbol\Sigma_c$ be a positive definite estimate of the covariance of the projected training features of $c\in\cCf$, and let $\bar{\boldsymbol\Sigma}=\frac{1}{K}\sum\nolimits_{c\in\cCf}\boldsymbol\Sigma_c$ be their average.
ERASE gives every remaining class the covariance $\mathbf R\propto\bar{\boldsymbol\Sigma}^{\frac{K}{K+1}}$ with the trace of $\bar{\boldsymbol\Sigma}$.
Its power comes from the posterior mean of a Gaussian model of log-covariances whose isotropic prior counts as one more class, so fewer forget classes give a rounder shape.
With one forget class, $\mathbf R$ is proportional to $\bar{\boldsymbol\Sigma}^{\frac12}$, the geometric mean of its shape and the identity.

\subsection{Transporting the Source Class}
Given the location $\boldsymbol\theta_k$ and shape $\mathbf R$ of the target, ERASE maps the offset of each selected feature from the mean of its source class $c$ by the matrix
\begin{equation}
\label{eq:transport}
\boldsymbol\Gamma_c=\boldsymbol\Sigma_c^{-\frac12}\bigl(\boldsymbol\Sigma_c^{\frac12}\mathbf R\,\boldsymbol\Sigma_c^{\frac12}\bigr)^{\frac12}\boldsymbol\Sigma_c^{-\frac12}.
\end{equation}
This matrix satisfies $\boldsymbol\Gamma_c\boldsymbol\Sigma_c\boldsymbol\Gamma_c=\mathbf R$.
For Gaussian classes, it gives the optimal transport between the two shapes with the smallest mean squared displacement~\cite{olkin1982distance}.
Being symmetric positive definite, it keeps projected features of one source and target distinct, as new tasks need.
Under conditions in the appendix, the exact final step also keeps them distinct.

For a selected input with source $c=c(\vz)$ and target $k$, write $\boldsymbol\eta_c(\vz)=\mP\vz-\mP\vmu_c$. ERASE gives it the feature
\begin{equation}
\label{eq:move}
\vu=\boldsymbol\theta_k+\boldsymbol\Gamma_c\boldsymbol\eta_c(\vz).
\end{equation}
\Cref{alg:erase} writes it as $\mathcal G_{ck}(\vu,q)$, where $q$ matters only for mixtures.
Groups that one map sends to $k$ take mean $\boldsymbol\theta_k$ and covariance $\mathbf R$ when their moments equal those the map uses (proof in the appendix).
Every map sends the fitted Gaussian of its source onto the same target Gaussian, the one distribution per target required by the second part of the recovery bound (\cref{prop:recovery}).

\subsection{Stored Target Distribution (ERASE-M)}
\label{sec:stored}
Instead of predicting each target distribution, \erasem{} takes it from statistics stored before a deletion request.
These are the count $N_j$, the projected mean $\mP\vmu_j$, and a covariance estimate $\boldsymbol\Sigma_j$ of the projected training features of every class $j$.
From source $c$ to target $k$, \erasem{} applies the full move of \cref{eq:move} with $\boldsymbol\theta_k=\mP\vmu_k$ and, in place of $\boldsymbol\Gamma_c$, the matrix $\boldsymbol\Gamma_{ck}$ of \cref{eq:transport} with $\boldsymbol\Sigma_k$ in place of $\mathbf R$, which satisfies $\boldsymbol\Gamma_{ck}\boldsymbol\Sigma_c\boldsymbol\Gamma_{ck}=\boldsymbol\Sigma_k$.
A group whose moments equal the source moments of its map then takes the stored class-$k$ moments, and the map sends the fitted Gaussian of the source onto that of class $k$.
A final step shared with the class-$k$ inputs keeps this match.
For a class with enough training samples, the service stores a Gaussian mixture instead, and \erasem{} uses the Gaussian mixture transport of~\cite{delon2020wasserstein}.

\begin{figure}[t]
\centering
\includegraphics{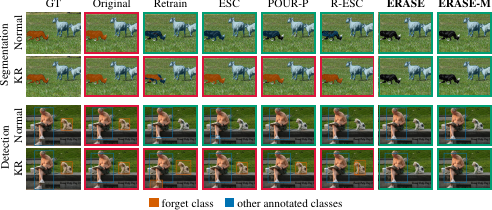}
\caption{\textbf{Normal and KR predictions with the class dog forgotten.} DeepLabv3+ and Faster R-CNN on PASCAL VOC 2012.
Red frames mark predictions that contain the dog.}
\label{fig:qualitative}
\end{figure}
\subsection{Completing the Edit}

The transport uses the target location and shape, but the head might not yet predict the target.
ERASE and \erasem{} therefore take the smallest nonnegative step $\beta^\star$ along $\vw_k$ giving $p_k\geq1-\delta$.
In classification, every input takes this step, and an unselected one starts with $\vu=\mP\vz$ and $k=F(\vx)$.
Every head row of the experiments has a larger inner product with itself than with any other row, so this step exists for every input (Appendix).
The head then keeps every correct remaining-class prediction and predicts the target of every edited feature, as \cref{prop:recovery} assumes (proof in the appendix).
If the class-$k$ inputs and the forget-class inputs with target $k$ all take this same step, it preserves the recovery bound (Appendix).
\label{sec:protect}
The final step also moves inputs of low confidence.
An unselected one far from every forget class takes a scaled step that keeps its prediction, and the power $\zeta$ keeps the order of confidences that flag errors and new-task inputs.

ERASE and \erasem{} treat each pixel or region in segmentation and detection as a query and also select it when a forget-class probability exceeds $\frac{1}{C}$.
Deleting a class makes its pixels and objects background, so a selected query takes its most probable label once the forget classes count as background.
A query labeled background becomes the background template, and in detection any other selected query keeps its score from the deployed detector.
Detection needs both the template and the moves (Appendix).

\section{Experiments}
\subsection{Experimental Setup}

\begin{table}[t]
\centering
\footnotesize
\renewcommand{\arraystretch}{0.88}
\setlength{\tabcolsep}{2pt}
\caption{\textbf{Classification with a frozen backbone under KR.} Test accuracy (\%) with $K$ forget classes on \mbox{VGGFace2-100} with ArcFace \mbox{ResNet-50}.
Parameter-based methods refit only the head of a frozen backbone, so their KR equals Original's.}
\label{tab:frozen-kr}
\resizebox{\linewidth}{!}{\begin{tabular}{@{}lc rr rr rr@{}}
\toprule
 & & \multicolumn{2}{c}{$K=1$} & \multicolumn{2}{c}{$K=5$} & \multicolumn{2}{c}{$K=20$} \\
\cmidrule(lr){3-4}\cmidrule(lr){5-6}\cmidrule(lr){7-8}
Method & $\mathcal C_r$ & $D_{ft}$$\downarrow$ & $\HMt$$\uparrow$ & $D_{ft}$$\downarrow$ & $\HMt$$\uparrow$ & $D_{ft}$$\downarrow$ & $\HMt$$\uparrow$ \\
\midrule
\multicolumn{8}{@{}l}{\textit{VGGFace2-100 / ArcFace ResNet-50}} \\
Original & \multirow{2}{*}{-} & 98.89\pmstd{\hz1.92} & \hz2.15\pmstd{\hz3.72} & 99.38\pmstd{0.56} & \hz1.23\pmstd{1.10} & 98.99\pmstd{0.21} & \hz2.00\pmstd{0.42} \\
Retrain & & 98.89\pmstd{\hz1.92} & \hz2.15\pmstd{\hz3.72} & 99.38\pmstd{0.56} & \hz1.23\pmstd{1.10} & 98.99\pmstd{0.21} & \hz2.00\pmstd{0.42} \\
\midrule
R-ESC & \multirow{2}{*}{\ding{51}} & \hz\textbf{0.00\pmstd{\hz0.00}} & \underline{99.68\pmstd{\hz0.01}} & \hz\textbf{0.00\pmstd{0.00}} & 99.67\pmstd{0.04} & \hz\textbf{0.00\pmstd{0.00}} & 99.64\pmstd{0.05} \\
LOFT & & \underline{43.33\pmstd{31.88}} & 67.96\pmstd{29.58} & \hz\underline{0.13\pmstd{0.13}} & 99.17\pmstd{0.03} & \hz0.42\pmstd{0.25} & 99.52\pmstd{0.15} \\
\midrule
ESC & \multirow{2}{*}{\ding{55}} & \hz\textbf{0.00\pmstd{\hz0.00}} & 99.66\pmstd{\hz0.01} & \hz\textbf{0.00\pmstd{0.00}} & 99.65\pmstd{0.06} & \hz\underline{0.03\pmstd{0.03}} & 99.70\pmstd{0.05} \\
POUR-P & & \hz\textbf{0.00\pmstd{\hz0.00}} & 99.67\pmstd{\hz0.00} & \hz\textbf{0.00\pmstd{0.00}} & \underline{99.68\pmstd{0.02}} & \hz\textbf{0.00\pmstd{0.00}} & \underline{99.74\pmstd{0.03}} \\
\rowcolor{eraserow}\textbf{ERASE} & \ding{55} & \hz\textbf{0.00\pmstd{\hz0.00}} & \textbf{99.83\pmstd{\hz0.01}} & \hz\textbf{0.00\pmstd{0.00}} & \textbf{99.82\pmstd{0.01}} & \hz\textbf{0.00\pmstd{0.00}} & \textbf{99.81\pmstd{0.04}} \\
\rowcolor{erasemrow}\textbf{ERASE-M} & \ding{55} & \hz\textbf{0.00\pmstd{\hz0.00}} & \textbf{99.83\pmstd{\hz0.01}} & \hz\textbf{0.00\pmstd{0.00}} & \textbf{99.82\pmstd{0.01}} & \hz\textbf{0.00\pmstd{0.00}} & \textbf{99.81\pmstd{0.04}} \\
\bottomrule
\end{tabular}}
\end{table}
\begin{table}[t]
\centering
\footnotesize
\renewcommand{\arraystretch}{0.88}
\setlength{\tabcolsep}{2pt}
\caption{\textbf{Segmentation and detection under KR.} IoU and AP at IoU 0.5 (\%) over three probe seeds. Marks as in \cref{tab:main-classification}.}
\label{tab:main-dense}
\resizebox{\linewidth}{!}{\begin{tabular}{@{}lc rr rr rr@{}}
\toprule
\multicolumn{8}{@{}l}{\textit{PASCAL VOC 2012 / DeepLabv3+}} \\
 & & \multicolumn{2}{c}{$K=1$} & \multicolumn{2}{c}{$K=5$} & \multicolumn{2}{c}{$K=10$} \\
\cmidrule(lr){3-4}\cmidrule(lr){5-6}\cmidrule(lr){7-8}
Method & $\mathcal C_r$ & IoU$_{ft}$$\downarrow$ & $\HMt$$\uparrow$ & IoU$_{ft}$$\downarrow$ & $\HMt$$\uparrow$ & IoU$_{ft}$$\downarrow$ & $\HMt$$\uparrow$ \\
\midrule
Original & \multirow{2}{*}{-} & 78.65\pmstd{0.40} & 32.19\pmstd{0.44} & 67.56\pmstd{0.20} & 43.44\pmstd{0.15} & 70.55\pmstd{0.40} & 39.91\pmstd{0.36} \\
Retrain & & 72.85\pmstd{1.09} & 38.60\pmstd{1.11} & 42.26\pmstd{2.74} & 61.61\pmstd{1.23} & 31.51\pmstd{0.94} & 64.98\pmstd{0.41} \\
\midrule
R-ESC & \ding{51} & \underline{49.72\pmstd{1.72}} & \underline{55.82\pmstd{1.09}} & \hz\underline{9.81\pmstd{4.74}} & 68.02\pmstd{2.61} & \underline{11.26\pmstd{1.32}} & 51.89\pmstd{7.89} \\
\midrule
ESC & \multirow{2}{*}{\ding{55}} & 78.48\pmstd{0.15} & 32.30\pmstd{0.15} & 11.38\pmstd{1.15} & \underline{74.73\pmstd{0.40}} & 17.47\pmstd{2.21} & 70.01\pmstd{0.90} \\
POUR-P & & 78.15\pmstd{0.49} & 32.77\pmstd{0.54} & 66.27\pmstd{0.27} & 44.59\pmstd{0.21} & 69.19\pmstd{0.27} & 41.13\pmstd{0.22} \\
\rowcolor{eraserow}\textbf{ERASE} & \ding{55} & \hz\textbf{0.00\pmstd{0.00}} & \textbf{79.82\pmstd{0.06}} & \hz\textbf{0.00\pmstd{0.00}} & \textbf{79.88\pmstd{0.08}} & \hz\textbf{0.00\pmstd{0.00}} & \underline{76.91\pmstd{0.03}} \\
\rowcolor{erasemrow}\textbf{ERASE-M} & \ding{55} & \hz\textbf{0.00\pmstd{0.00}} & \textbf{79.82\pmstd{0.06}} & \hz\textbf{0.00\pmstd{0.00}} & \textbf{79.88\pmstd{0.08}} & \hz\textbf{0.00\pmstd{0.00}} & \textbf{76.92\pmstd{0.03}} \\
\midrule
\multicolumn{8}{@{}l}{\textit{PASCAL VOC 2012 / Faster R-CNN}} \\
Method & $\mathcal C_r$ & AP$_{ft}$$\downarrow$ & $\HMt$$\uparrow$ & AP$_{ft}$$\downarrow$ & $\HMt$$\uparrow$ & AP$_{ft}$$\downarrow$ & $\HMt$$\uparrow$ \\
\midrule
Original & \multirow{2}{*}{-} & 71.86\pmstd{1.28} & 37.50\pmstd{1.07} & 63.38\pmstd{0.68} & 44.00\pmstd{0.40} & 60.92\pmstd{0.57} & 45.22\pmstd{0.28} \\
Retrain & & 60.78\pmstd{1.16} & 45.94\pmstd{0.73} & 40.11\pmstd{0.57} & 56.31\pmstd{0.36} & 28.04\pmstd{0.98} & 60.26\pmstd{0.35} \\
\midrule
R-ESC & \ding{51} & 60.66\pmstd{4.74} & 45.28\pmstd{2.91} & 55.10\pmstd{2.14} & 45.07\pmstd{1.10} & 51.85\pmstd{0.91} & 35.46\pmstd{0.76} \\
\midrule
ESC & \multirow{2}{*}{\ding{55}} & 68.03\pmstd{1.87} & 40.46\pmstd{1.50} & 50.84\pmstd{0.11} & 51.01\pmstd{0.03} & 58.37\pmstd{0.44} & \underline{46.03\pmstd{0.22}} \\
POUR-P & & 71.83\pmstd{1.56} & 37.51\pmstd{1.31} & 62.64\pmstd{0.90} & 44.48\pmstd{0.52} & 60.06\pmstd{0.34} & 45.69\pmstd{0.15} \\
\rowcolor{eraserow}\textbf{ERASE} & \ding{55} & \hz\textbf{0.32\pmstd{0.05}} & \underline{68.75\pmstd{0.03}} & \hz\textbf{0.25\pmstd{0.04}} & \underline{68.02\pmstd{0.11}} & \hz\textbf{0.13\pmstd{0.02}} & \textbf{67.07\pmstd{0.14}} \\
\rowcolor{erasemrow}\textbf{ERASE-M} & \ding{55} & \hz\underline{0.35\pmstd{0.14}} & \textbf{68.88\pmstd{0.01}} & \hz\underline{0.49\pmstd{0.06}} & \textbf{68.06\pmstd{0.03}} & \hz\underline{0.24\pmstd{0.02}} & \textbf{67.07\pmstd{0.16}} \\
\midrule
\multicolumn{8}{@{}l}{\textit{MS COCO / Faster R-CNN}} \\
 & & \multicolumn{2}{c}{$K=1$} & \multicolumn{2}{c}{$K=5$} & \multicolumn{2}{c}{$K=20$} \\
\cmidrule(lr){3-4}\cmidrule(lr){5-6}\cmidrule(lr){7-8}
Original & \multirow{2}{*}{-} & 40.41\pmstd{0.23} & 51.48\pmstd{0.20} & 34.08\pmstd{0.54} & 54.22\pmstd{0.17} & 45.57\pmstd{0.25} & 49.51\pmstd{0.25} \\
Retrain & & 16.32\pmstd{0.99} & 57.74\pmstd{0.88} & 19.72\pmstd{0.80} & 57.09\pmstd{0.67} & 23.61\pmstd{0.40} & 55.54\pmstd{0.69} \\
\midrule
R-ESC & \ding{51} & \underline{27.55\pmstd{1.83}} & 54.79\pmstd{0.50} & 30.34\pmstd{1.46} & 54.52\pmstd{0.62} & 40.21\pmstd{0.97} & 45.45\pmstd{0.37} \\
\midrule
ESC & \multirow{2}{*}{\ding{55}} & 37.77\pmstd{1.75} & 52.21\pmstd{0.68} & 28.02\pmstd{0.72} & 55.44\pmstd{0.31} & 43.01\pmstd{0.53} & 49.47\pmstd{0.30} \\
POUR-P & & 38.86\pmstd{0.57} & 52.03\pmstd{0.35} & 35.13\pmstd{0.61} & 53.80\pmstd{0.56} & 43.33\pmstd{0.31} & 50.10\pmstd{0.32} \\
\rowcolor{eraserow}\textbf{ERASE} & \ding{55} & \hz\textbf{0.00\pmstd{0.00}} & \underline{61.87\pmstd{0.30}} & \hz\underline{0.50\pmstd{0.13}} & \textbf{62.00\pmstd{0.20}} & \hz\underline{0.45\pmstd{0.15}} & \textbf{61.21\pmstd{0.14}} \\
\rowcolor{erasemrow}\textbf{ERASE-M} & \ding{55} & \hz\textbf{0.00\pmstd{0.00}} & \textbf{61.90\pmstd{0.22}} & \hz\textbf{0.32\pmstd{0.09}} & \underline{61.60\pmstd{0.48}} & \hz\textbf{0.21\pmstd{0.06}} & \underline{59.17\pmstd{0.42}} \\
\bottomrule
\end{tabular}}
\end{table}

\begin{figure}[t]
\centering
\includegraphics[width=\columnwidth]{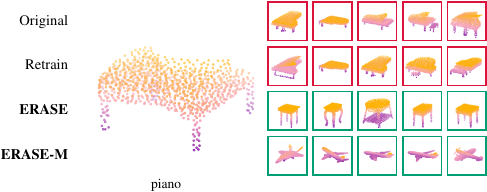}\\[3pt]
\includegraphics[width=\columnwidth]{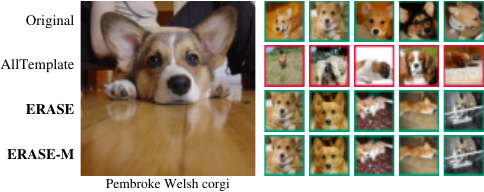}
\caption{\textbf{Retrieval with edited features.} Five nearest training items at $K=1$ of a forget-class piano on ModelNet40 with DGCNN (top) and of an image from a new ImageNet class on \mbox{Tiny-ImageNet} with CLIP \mbox{ViT-L/14} (bottom).
Red frames mark pianos (top), and green frames mark the query class (bottom).}
\label{fig:pointcloud}
\end{figure}
\paragraph{Datasets and models.} Classification uses \mbox{ResNet-18}~\cite{he2016deep} on \mbox{CIFAR-100}~\cite{krizhevsky2009learning} and \mbox{Tiny-ImageNet}~\cite{le2015tiny} and DGCNN~\cite{wang2019dynamic} on ModelNet40~\cite{wu20153d} point clouds.
Frozen backbones with linear heads classify \mbox{VGGFace2-100}~\cite{cao2018vggface2} with ArcFace \mbox{ResNet-50}~\cite{deng2019arcface}, \mbox{CIFAR-100} and \mbox{Tiny-ImageNet} with CLIP \mbox{ViT-L/14}~\cite{radford2021learning}, and \mbox{Market-1501}~\cite{zheng2015scalable} with DINOv2 \mbox{ViT-L/14}~\cite{oquab2023dinov2}.
Segmentation uses DeepLabv3+~\cite{chen2018encoder} on PASCAL VOC 2012~\cite{everingham2010pascal}, and detection uses Faster R-CNN~\cite{ren2015faster} on PASCAL VOC 2012 and MS COCO~\cite{lin2014microsoft}.
Classification has 21 settings, three values of $K$ per model, each with three forget sets, and segmentation and detection have one in the main tables and three in the appendix.

\paragraph{Baselines.}We compare with the parameter-based methods Finetune~\cite{golatkar2020eternal}, NG~\cite{thudi2022unrolling}, RL~\cite{graves2021amnesiac}, SalUn~\cite{fan2024salun}, BadT~\cite{chundawat2023can}, and DELETE~\cite{zhou2025decoupled} and with the feature-level methods ESC~\cite{lee2025esc}, \pourp~\cite{le2026pour}, \resc~\cite{jeon2026r}, and LOFT~\cite{fang2026machine}.
Every baseline keeps its published algorithm and searches the hyperparameters of its paper.

\paragraph{Evaluation metrics.}Normal applies the deployed head to $\tilde\vz$, and KR trains a linear probe on the training $\tilde\vz$ with the labels of all classes, following ESC~\cite{lee2025esc} and \resc~\cite{jeon2026r}.
We report the accuracies $D_{ft}$ and $D_{rt}$ on forget-class and remaining-class test samples and $\HMt$, the harmonic mean of $100-D_{ft}$ and $D_{rt}$.
An MLP tests nonlinear recovery, and segmentation and detection report IoU and AP alike.
Classification tables give the mean $\pm$ standard deviation over three forget sets.

\paragraph{Implementation details.} ERASE and \erasem{} fix $\epsilon=10^{-5}$, the power $\zeta=\frac{1}{32}$ of \cref{alg:erase}, and the threshold of the scaled step (Appendix) for all settings.
Every other method takes its configuration, and every probe its hyperparameters, by one rule on a selection half of each test set.
All results come from the other half, and swapping the halves keeps the highest KR $\HMt$ with ERASE or \erasem{} in every column of \cref{tab:main-classification,tab:main-dense}.
We use eight NVIDIA RTX A6000 GPUs.

\subsection{Recovery by New Probes}

In classification under Normal, ERASE removes the forget classes from the deployed predictions and keeps Original's remaining-class accuracy, while several baselines still predict some forget-class inputs correctly (\cref{tab:main-classification}).
Under KR, Retrain and the parameter-based methods either keep the forget classes recognizable or lose the remaining classes too.
ERASE and \erasem{} keep forget-class accuracy near zero with the highest $\HMt$ in every column, also with frozen backbones (\cref{tab:frozen-kr}).
In segmentation and detection, KR still recovers the forget classes from every baseline.
ERASE and \erasem{} keep IoU$_{ft}$ and AP$_{ft}$ near zero with a higher $\HMt$ than every baseline in nearly every column, also on two more forget sets (\cref{tab:main-dense} and Appendix).

In \cref{fig:qualitative}, KR recovers the dog from the features of Retrain, ESC, \pourp, and \resc{} but not of ERASE and \erasem, where it still recognizes the other objects.
In \cref{fig:pointcloud}, the nearest training shapes of a piano are all pianos for Retrain and none for ERASE.

\begin{figure}[t]
\centering
\includegraphics[width=\columnwidth]{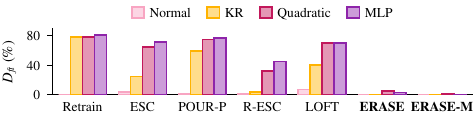}
\caption{\textbf{Recovery by the order of the probe.} $D_{ft}$ (\%, $\downarrow$) under each probe, averaged over the 21 settings.}
\label{fig:order}
\end{figure}

From ESC, \resc, and LOFT, the MLP recovers more of the forget classes than KR, and from \pourp{} both probes recover most of them (\cref{fig:order}).
Under membership inference, ERASE and \erasem{} score lowest among unlearning methods in every classification setting (Appendix).

\begin{table}[t]
\centering
\small
\renewcommand{\arraystretch}{0.88}
\setlength{\tabcolsep}{2.5pt}
\caption{\textbf{Targeted probes.} AUC (\%) of binary and target probes and forget accuracy (\%) on \mbox{CIFAR-100} with \mbox{ResNet-18}.
Bold and underline mark values closest to chance, and \ding{51} is as in \cref{tab:main-classification}.}
\label{tab:targeted}
\resizebox{\linewidth}{!}{\begin{tabular}{@{}lcrrrrrrrr@{}}
\toprule
Method & $\mathcal C_r$ & \multicolumn{3}{c}{Binary} & \multicolumn{3}{c}{Target} & \multicolumn{2}{c}{Forget} \\
\cmidrule(lr){3-5}\cmidrule(lr){6-8}\cmidrule(lr){9-10}
 & & $K=1$ & $K=5$ & $K=20$ & $K=1$ & $K=5$ & $K=20$ & $K=5$ & $K=20$ \\
\midrule
Original & \multirow{2}{*}{-} & 96.1\pmstd{3.1} & 97.1\pmstd{1.7} & 96.4\pmstd{0.7} & 92.0\pmstd{\hz8.0} & 96.6\pmstd{2.1} & 95.8\pmstd{0.1} & 97.0\pmstd{1.3} & 91.7\pmstd{1.5} \\
Retrain & & 98.4\pmstd{1.3} & 95.3\pmstd{2.6} & 89.8\pmstd{1.2} & 89.1\pmstd{11.7} & 92.2\pmstd{4.3} & 89.7\pmstd{1.0} & 92.9\pmstd{2.5} & 79.7\pmstd{1.2} \\
\midrule
R-ESC & \multirow{2}{*}{\ding{51}} & 94.9\pmstd{3.8} & 94.4\pmstd{2.7} & \underline{79.4\pmstd{1.6}} & 83.2\pmstd{10.3} & \underline{91.1\pmstd{2.7}} & \underline{81.0\pmstd{0.4}} & 85.0\pmstd{6.3} & 45.7\pmstd{2.0} \\
LOFT & & 97.9\pmstd{1.5} & 96.6\pmstd{2.0} & 94.9\pmstd{0.8} & 99.6\pmstd{\hz0.4} & 96.4\pmstd{2.3} & 95.5\pmstd{0.8} & 87.0\pmstd{6.6} & 78.3\pmstd{2.5} \\
\midrule
ESC & \multirow{2}{*}{\ding{55}} & 94.9\pmstd{4.1} & 96.6\pmstd{1.8} & 94.9\pmstd{0.8} & 93.6\pmstd{\hz6.1} & 96.7\pmstd{2.0} & 95.9\pmstd{0.8} & 89.1\pmstd{2.6} & 77.3\pmstd{1.6} \\
POUR-P & & 96.7\pmstd{2.4} & 97.2\pmstd{1.7} & 95.7\pmstd{0.5} & 93.7\pmstd{\hz6.0} & 96.5\pmstd{2.3} & 95.9\pmstd{0.1} & 94.1\pmstd{0.7} & 87.2\pmstd{1.5} \\
\rowcolor{eraserow}\textbf{ERASE} & \ding{55} & \underline{80.5\pmstd{3.6}} & \underline{88.5\pmstd{5.6}} & 90.4\pmstd{0.6} & \underline{81.0\pmstd{\hz4.0}} & 91.7\pmstd{4.0} & 94.2\pmstd{0.3} & \underline{23.3\pmstd{2.0}} & \hz\underline{5.8\pmstd{0.2}} \\
\rowcolor{erasemrow}\textbf{ERASE-M} & \ding{55} & \textbf{53.0\pmstd{1.4}} & \textbf{51.7\pmstd{1.3}} & \textbf{50.4\pmstd{0.4}} & \textbf{53.1\pmstd{\hz2.1}} & \textbf{51.9\pmstd{1.8}} & \textbf{51.0\pmstd{0.7}} & \textbf{22.3\pmstd{0.9}} & \hz\textbf{5.0\pmstd{0.2}} \\
\bottomrule
\end{tabular}}
\end{table}
\begin{table}[t]
\centering
\small
\renewcommand{\arraystretch}{0.88}
\setlength{\tabcolsep}{3pt}
\caption{\textbf{Detectors of increasing capacity.} AUC (\%) of the probes of \cref{tab:targeted} as linear, RBF-kernel, and one- or two-hidden-layer MLP detectors, averaged over the 21 settings.
Marks as in \cref{tab:targeted}.}
\label{tab:capacity}
\resizebox{\linewidth}{!}{\begin{tabular}{@{}lcrrrrrrr@{}}
\toprule
Method & $\mathcal C_r$ & \multicolumn{4}{c}{Binary} & \multicolumn{3}{c}{Target} \\
\cmidrule(lr){3-6}\cmidrule(lr){7-9}
 & & Lin. & RBF & MLP$_1$ & MLP$_2$ & Lin. & MLP$_1$ & MLP$_2$ \\
\midrule
Original & \multirow{2}{*}{-} & 94.9\pmstd{0.7} & 94.2\pmstd{0.5} & 96.7\pmstd{0.5} & 96.5\pmstd{0.6} & 91.3\pmstd{1.0} & 94.8\pmstd{1.1} & 94.7\pmstd{1.0} \\
Retrain & & 93.1\pmstd{0.8} & 93.2\pmstd{0.6} & 95.8\pmstd{0.4} & 95.7\pmstd{0.5} & 87.5\pmstd{1.1} & 92.9\pmstd{0.7} & 93.1\pmstd{0.6} \\
\midrule
R-ESC & \multirow{2}{*}{\ding{51}} & 67.7\pmstd{0.9} & 75.0\pmstd{1.1} & 89.8\pmstd{1.2} & 89.1\pmstd{1.4} & \underline{54.7\pmstd{1.1}} & 87.4\pmstd{0.8} & 86.8\pmstd{1.2} \\
LOFT & & 91.8\pmstd{1.0} & 94.5\pmstd{0.2} & 95.5\pmstd{0.5} & 95.4\pmstd{0.3} & 92.1\pmstd{1.2} & 96.0\pmstd{0.5} & 96.0\pmstd{0.2} \\
\midrule
ESC & \multirow{2}{*}{\ding{55}} & 92.4\pmstd{0.4} & 93.0\pmstd{0.6} & 94.8\pmstd{0.6} & 94.3\pmstd{0.7} & 92.0\pmstd{0.8} & 94.9\pmstd{0.5} & 94.6\pmstd{0.3} \\
POUR-P & & 90.0\pmstd{0.3} & 93.2\pmstd{0.3} & 95.3\pmstd{0.3} & 95.1\pmstd{0.4} & 87.6\pmstd{0.9} & 94.1\pmstd{1.1} & 94.2\pmstd{1.3} \\
\rowcolor{eraserow}\textbf{ERASE} & \ding{55} & 55.8\pmstd{1.3} & 66.0\pmstd{0.6} & 76.0\pmstd{0.6} & 76.8\pmstd{0.9} & 55.3\pmstd{1.0} & 79.1\pmstd{1.3} & 79.5\pmstd{0.6} \\
\rowcolor{erasemrow}\textbf{ERASE-M} & \ding{55} & \textbf{54.1\pmstd{0.9}} & \textbf{54.0\pmstd{0.8}} & \textbf{54.4\pmstd{1.2}} & \textbf{54.1\pmstd{0.6}} & \textbf{54.0\pmstd{1.0}} & \textbf{53.9\pmstd{0.5}} & \textbf{54.1\pmstd{0.4}} \\
ERASE-M ($\boldsymbol\Gamma=\mathbf I$) & \ding{55} & \underline{54.6\pmstd{1.3}} & \underline{58.4\pmstd{0.9}} & \underline{71.8\pmstd{1.6}} & \underline{73.8\pmstd{1.7}} & 55.5\pmstd{1.0} & \underline{73.6\pmstd{2.2}} & \underline{76.2\pmstd{2.3}} \\
\bottomrule
\end{tabular}}
\end{table}
\begin{table}[t]
\centering
\small
\renewcommand{\arraystretch}{0.88}
\setlength{\tabcolsep}{4pt}
\caption{\textbf{Transport targets.} Forget-class accuracy (\%) of KR and the MLP and AUC (\%) of the binary and target probes with MLP$_2$, averaged over the 21 settings.
Pooled Gaussian reads the statistics of \erasem{} and moves the selected inputs to one Gaussian of all remaining classes.
Bold and underline mark the best two unlearning methods, the AUC by its distance to chance.}
\label{tab:transport-main}
\begin{tabular}{@{}l rr rr@{}}
\toprule
 & KR & MLP & \multicolumn{2}{c}{AUC} \\
\cmidrule(lr){4-5}
Method & $D_{ft}$$\downarrow$ & $D_{ft}$$\downarrow$ & Binary & Target \\
\midrule
Original & 80.11\pmstd{2.17} & 82.65\pmstd{2.62} & 96.46\pmstd{0.55} & 94.72\pmstd{1.01} \\
Retrain & 78.44\pmstd{3.06} & 80.94\pmstd{3.33} & 95.66\pmstd{0.47} & 93.05\pmstd{0.60} \\
\midrule
LEACE~\cite{belrose2306leace} & 22.83\pmstd{1.95} & 71.01\pmstd{2.65} & 94.74\pmstd{0.70} & 93.58\pmstd{0.64} \\
QLEACE~\cite{quirke2025slowing} & \hz\underline{0.11\pmstd{0.10}} & \hz9.27\pmstd{2.18} & \underline{83.15\pmstd{0.57}} & 85.64\pmstd{0.95} \\
Pooled Gaussian & \hz0.13\pmstd{0.14} & \hz\underline{9.10\pmstd{1.74}} & 83.40\pmstd{0.68} & \underline{85.22\pmstd{1.53}} \\
\midrule
\rowcolor{erasemrow}\textbf{ERASE-M} & \hz\textbf{0.04\pmstd{0.05}} & \hz\textbf{0.09\pmstd{0.03}} & \textbf{54.12\pmstd{0.62}} & \textbf{54.13\pmstd{0.43}} \\
\bottomrule
\end{tabular}
\end{table}
\subsection{Distinguishability}
A recipient with forget-class labels can also train three targeted probes (\cref{tab:targeted}).
Besides the binary probe of \cref{def:recovery}, a target probe separates forget-class inputs with target $k$ from class-$k$ inputs, and a forget probe names their forget class.
\erasem{} keeps all three probes near chance (AUC 50 and forget accuracy $\frac{100}{K}$) for every $K$.
Even after a perfect detector, only ERASE and \erasem{} keep the forget probe near chance, unlike Retrain and every baseline.
In the other settings, no baseline has a lower target AUC than \erasem{} (Appendix).
On \mbox{VGGFace2-100} with ArcFace \mbox{ResNet-50}, KR recovers almost nothing from ESC, \pourp, \resc, or ERASE (\cref{tab:frozen-kr}), yet the target probe separates every baseline's edits (Appendix).

Linear detectors barely separate the moved inputs of ERASE and \erasem.
\erasem{} stays closest to chance for every nonlinear detector, nearest neighbors, gradient-boosted trees, detectors of higher moments, and adaptive detectors included, also at low false-positive rates (\cref{tab:capacity} and Appendix).
With the identity transport ($\boldsymbol\Gamma=\mathbf I$), \erasem{} replaces the source center by the target center, and MLP detectors separate its moved inputs.
They also separate inputs transported to one Gaussian of all remaining classes or to the barycenter of QLEACE~\cite{quirke2025slowing}, while the binary probe stays near chance for \erasem{} (\cref{tab:transport-main}).
\erasem{} needs both its stored location and its stored shape, since predicting either one lets the MLP separate its edits much better (Appendix).
With 32 eigenpairs per class in place of full covariances, it stays near chance (Appendix).

Adding the projection and the final step of ERASE to the feature-level baselines keeps their deployed predictions (Appendix).
The MLP still recovers far more of their forget classes than of those of ERASE, and the binary probe still separates their edits, unlike those of \erasem.

\subsection{Utility on New Tasks}
\label{sec:exp-utility}
\begin{figure}[t]
\centering
\includegraphics[width=\columnwidth]{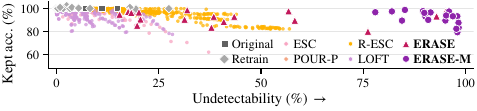}
\caption{\textbf{Utility and undetectability.} Kept accuracy (the new-task accuracy as a percentage of that of Original on its unedited features $g(\vx)$, on the tasks of \cref{sec:exp-utility}) against undetectability ($100-2|\mathrm{AUC}-50|$ for the binary MLP$_1$ probe).
Small markers give every searched configuration of each baseline for each model and $K$, and large markers the other methods.}
\label{fig:tradeoff}
\vspace{1pt}
\end{figure}

ERASE and \erasem{} keep most of the original accuracy on new identities, new ImageNet classes~\cite{deng2009imagenet}, and new ShapeNetPart classes~\cite{yi2016scalable} (\cref{fig:tradeoff}), and most of the retrieval mAP of new identities (Appendix).
AllTemplate gives each input the template $\vt_k$ of its target $k$ in ERASE, so the two share every deployed prediction and images with one target share one feature (\cref{fig:pointcloud}).
ERASE keeps far more accuracy than AllTemplate and exceeds the bound of \cref{sec:principle}, so the differences it keeps within each target carry information for new tasks (Appendix).
A draw from the target distribution also matches location and shape, but leaves the moved inputs nothing beyond their target (\cref{sec:principle}).
For the binary MLP$_1$ probe, no configuration that the feature-level baselines search comes as close to chance as \erasem, which keeps most of Original's new-task accuracy (\cref{fig:tradeoff}).

\begin{table}[t]
\centering
\small
\renewcommand{\arraystretch}{0.88}
\setlength{\tabcolsep}{3pt}
\caption{\textbf{Factors of ERASE and \erasem.} Accuracy (\%) over the 21 settings.
A check keeps a factor, from left to right the transport, the move of \cref{eq:move}, key targets, and the predicted location.
Bold and underline mark the best two values of each method.}
\label{tab:components}
\resizebox{\linewidth}{!}{\begin{tabular}{@{}lcccc rrr rrr@{}}
\toprule
 & & & & & \multicolumn{3}{c}{KR} & \multicolumn{3}{c}{MLP} \\
\cmidrule(lr){6-8}\cmidrule(lr){9-11}
Method\rule{0pt}{30pt} & \smash{\rotatebox{90}{Transport}} & \smash{\rotatebox{90}{Full move}} & \smash{\rotatebox{90}{Key target}} & \smash{\rotatebox{90}{Location}} & $D_{ft}$$\downarrow$ & $D_{rt}$$\uparrow$ & $\HMt$$\uparrow$ & $D_{ft}$$\downarrow$ & $D_{rt}$$\uparrow$ & $\HMt$$\uparrow$ \\
\midrule
 & & \ding{51} & \ding{51} & \ding{51} & \hz0.08\pmstd{0.07} & \underline{81.29\pmstd{0.16}} & \underline{89.08\pmstd{0.06}} & 23.78\pmstd{2.07} & 80.70\pmstd{0.16} & 76.49\pmstd{1.66} \\
 & \ding{51} & & \ding{51} & \ding{51} & \hz\textbf{0.01\pmstd{0.01}} & 81.28\pmstd{0.16} & \textbf{89.10\pmstd{0.09}} & \hz\textbf{1.99\pmstd{0.70}} & \textbf{81.14\pmstd{0.16}} & \textbf{88.16\pmstd{0.34}} \\
 & \ding{51} & \ding{51} & & \ding{51} & 17.05\pmstd{0.86} & \textbf{81.34\pmstd{0.25}} & 80.84\pmstd{0.48} & 37.96\pmstd{4.66} & 80.83\pmstd{0.20} & 67.30\pmstd{3.52} \\
 & \ding{51} & \ding{51} & \ding{51} & & \hz0.11\pmstd{0.05} & \underline{81.29\pmstd{0.15}} & 89.07\pmstd{0.07} & \hz7.64\pmstd{2.07} & \underline{81.04\pmstd{0.17}} & 85.28\pmstd{1.20} \\
\midrule
\rowcolor{eraserow}\textbf{ERASE} & \ding{51} & \ding{51} & \ding{51} & \ding{51} & \hz\underline{0.05\pmstd{0.05}} & 81.28\pmstd{0.16} & \underline{89.08\pmstd{0.08}} & \hz\underline{3.01\pmstd{0.10}} & \textbf{81.14\pmstd{0.17}} & \underline{87.76\pmstd{0.17}} \\
\midrule
 & & \ding{51} & \ding{51} & \ding{51} & \hz\textbf{0.00\pmstd{0.00}} & \textbf{81.29\pmstd{0.16}} & \textbf{89.11\pmstd{0.09}} & 19.55\pmstd{2.24} & 80.87\pmstd{0.23} & 79.25\pmstd{1.51} \\
 & \ding{51} & \ding{51} & & \ding{51} & \hz6.27\pmstd{0.82} & 81.10\pmstd{0.25} & 86.25\pmstd{0.52} & \underline{11.94\pmstd{4.17}} & \underline{81.03\pmstd{0.48}} & \underline{83.46\pmstd{2.37}} \\
\midrule
\rowcolor{erasemrow}\textbf{ERASE-M} & \ding{51} & \ding{51} & \ding{51} & \ding{51} & \hz\underline{0.04\pmstd{0.05}} & \underline{81.28\pmstd{0.16}} & \underline{89.09\pmstd{0.11}} & \hz\textbf{0.09\pmstd{0.03}} & \textbf{81.26\pmstd{0.15}} & \textbf{89.05\pmstd{0.07}} \\
\bottomrule
\end{tabular}}
\vspace{1pt}
\end{table}

\subsection{Ablation and Speed}
\label{sec:exp-ablation}

Changing any one factor of ERASE keeps its remaining-class accuracy under KR (\cref{tab:components}).
Only targets ranked by the scores of the head, in place of key targets, raise the recovery of KR, since they depend on the source.
Replacing the transport or the location raises the recovery of the MLP, the identity transport the most, also for \erasem, which has the lowest recovery of the MLP. Setting $\alpha=1$, the other case of \cref{sec:replacement}, also raises the MLP recovery of ERASE (Appendix).
For the tested $\epsilon$ and $\zeta$ and powers of $\bar{\boldsymbol\Sigma}$ from $0$ to $\frac{K}{K+2}$, the recovery of ERASE stays low (Appendix).
The power $1$, which drops the isotropic prior, raises the recovery of the MLP. Without the projection, probes recover the forget classes, and dropping the ignored directions only from moved features lets a linear detector find them (Appendix).

The gate selects most forget-class test inputs, nearly all with their true class as source, and few remaining-class ones (Appendix).
KR and the MLP recover almost none of the forget-class inputs that the gate misses, which the head already assigns to remaining classes (Appendix).
With the fixed $\zeta$, the deployed head flags its errors and new-task inputs nearly as well as on the original features, whereas a zero power leaves these flags near chance (Appendix).

With the final step for moved inputs only, the MLP can recover an entire forget class, while the shared step keeps this recovery low (Appendix).
With one Gaussian per class in place of the stored mixtures, the binary and target probes separate the edits of \erasem{} better (Appendix).
Either method handles a deletion request orders of magnitude faster than retraining a trained classification network and edits a query in milliseconds (Appendix).

\section{Conclusion}
Once features are released, a deletion must hold for the classifiers that recipients train on them.
Retrained features keep the forget classes recognizable, more so for more accurate models.
Such a model already predicts its remaining-class distributions, and ERASE moves the forget classes onto them in location and in shape, as the recovery bound requires.
With stored moments, \erasem{} keeps detectors near chance.
New classifiers rarely recognize them.
Both methods also keep most of Original's new-task accuracy, so deleting a class from released features needs no retraining.

\clearpage
{
\small
\bibliographystyle{ieeenat_fullname}
\bibliography{main}

@String(CVPR= {IEEE Conf. Comput. Vis. Pattern Recog.})

@String(ICCV= {Int. Conf. Comput. Vis.})

@String(ECCV= {Eur. Conf. Comput. Vis.})

@String(TOG= {ACM Trans. Graph.})

@String(ICLR = {Int. Conf. Learn. Represent.})

@String(AAAI = {AAAI})

@String(CVPR  = {CVPR})

@String(ICCV  = {ICCV})

@String(ECCV  = {ECCV})

@String(TOG   = {ACM TOG})

@String(ICLR  = {ICLR})

@inproceedings{lee2025esc,
  author       = {Tae{-}Young Lee and
                  Sundong Park and
                  Minwoo Jeon and
                  Hyoseok Hwang and
                  Gyeong{-}Moon Park},
  title        = {{ESC:} Erasing Space Concept for Knowledge Deletion},
  booktitle    = {{IEEE/CVF} Conference on Computer Vision and Pattern Recognition,
                  {CVPR} 2025, Nashville, TN, USA, June 11-15, 2025},
  pages        = {5010--5019},
  publisher    = {Computer Vision Foundation / {IEEE}},
  year         = {2025},
  url          = {https://openaccess.thecvf.com/content/CVPR2025/html/Lee\_ESC\_Erasing\_Space\_Concept\_for\_Knowledge\_Deletion\_CVPR\_2025\_paper.html},
  doi          = {10.1109/CVPR52734.2025.00472},
  bibsource    = {dblp computer science bibliography, https://dblp.org}
}

@inproceedings{jeon2026r,
  author       = {Kang Eun Jeon and
                  Yunsung Kang and
                  Do Yeong Kang and
                  Tae{-}Young Lee and
                  Gyeong{-}Moon Park and
                  Jong Hwan Ko},
  editor       = {Paolo Favaro and
                  Zuzana Kukelova and
                  Atsuto Maki and
                  Anna Rohrbach and
                  Konrad Schindler and
                  Federico Tombari},
  title        = {{R-ESC:} Robustly Erasing Space Concepts via Stochastic Feature Remapping},
  booktitle    = {Computer Vision - {ECCV} 2026 - 19th European Conference, Malm{\"{o}},
                  Sweden, September 8-12, 2026, Proceedings, Part {XIX}},
  series       = {Lecture Notes in Computer Science},
  volume       = {17019},
  pages        = {635--652},
  publisher    = {Springer},
  year         = {2026},
  url          = {https://doi.org/10.1007/978-3-032-37550-6\_36},
  doi          = {10.1007/978-3-032-37550-6\_36},
  bibsource    = {dblp computer science bibliography, https://dblp.org}
}

@inproceedings{le2026pour,
  author       = {Anjie Le and
                  Can Peng and
                  Yuyuan Liu and
                  J. Alison Noble},
  title        = {{POUR:} {A} Provably Optimal Method for Unlearning Representation
                  via Neural Collapse},
  booktitle    = {{IEEE/CVF} Conference on Computer Vision and Pattern Recognition,
                  {CVPR} 2026, Denver, CO, USA, June 3-7, 2026},
  pages        = {10273--10282},
  publisher    = {Computer Vision Foundation},
  year         = {2026},
  url          = {https://openaccess.thecvf.com/content/CVPR2026/html/Le\_POUR\_A\_Provably\_Optimal\_Method\_for\_Unlearning\_Representation\_via\_Neural\_CVPR\_2026\_paper.html},
  bibsource    = {dblp computer science bibliography, https://dblp.org}
}

@article{fang2026machine,
  title={Machine Unlearning in Low-Dimensional Feature Subspace},
  author={Fang, Kun and Tao, Qinghua and Liu, Junxu and Xiao, Yaxin and Ye, Qingqing and Sun, Jian and Hu, Haibo},
  journal={arXiv preprint arXiv:2601.22456},
  year={2026}
}

@inproceedings{zhou2025decoupled,
  title={Decoupled distillation to erase: A general unlearning method for any class-centric tasks},
  author={Zhou, Yu and Zheng, Dian and Mo, Qijie and Lu, Renjie and Lin, Kun-Yu and Zheng, Wei-Shi},
  booktitle={2025 IEEE/CVF Conference on Computer Vision and Pattern Recognition (CVPR)},
  pages={20350--20359},
  year={2025},
  organization={IEEE}
}

@inproceedings{fan2024salun,
  author       = {Chongyu Fan and
                  Jiancheng Liu and
                  Yihua Zhang and
                  Eric Wong and
                  Dennis Wei and
                  Sijia Liu},
  title        = {SalUn: Empowering Machine Unlearning via Gradient-based Weight Saliency
                  in Both Image Classification and Generation},
  booktitle    = {The Twelfth International Conference on Learning Representations,
                  {ICLR} 2024, Vienna, Austria, May 7-11, 2024},
  publisher    = {OpenReview.net},
  year         = {2024},
  url          = {https://openreview.net/forum?id=gn0mIhQGNM},
  bibsource    = {dblp computer science bibliography, https://dblp.org}
}

@inproceedings{chundawat2023can,
  title={Can bad teaching induce forgetting? unlearning in deep networks using an incompetent teacher},
  author={Chundawat, Vikram S and Tarun, Ayush K and Mandal, Murari and Kankanhalli, Mohan},
  booktitle={Proceedings of the AAAI Conference on Artificial Intelligence},
  volume={37},
  number={6},
  pages={7210--7217},
  year={2023}
}

@inproceedings{golatkar2020eternal,
  title={Eternal sunshine of the spotless net: Selective forgetting in deep networks},
  author={Golatkar, Aditya and Achille, Alessandro and Soatto, Stefano},
  booktitle={2020 IEEE/CVF Conference on Computer Vision and Pattern Recognition (CVPR)},
  pages={9301--9309},
  year={2020},
  organization={IEEE}
}

@inproceedings{graves2021amnesiac,
  title={Amnesiac machine learning},
  author={Graves, Laura and Nagisetty, Vineel and Ganesh, Vijay},
  booktitle={Proceedings of the AAAI conference on artificial intelligence},
  volume={35},
  number={13},
  pages={11516--11524},
  year={2021}
}

@inproceedings{thudi2022unrolling,
  title={Unrolling sgd: Understanding factors influencing machine unlearning},
  author={Thudi, Anvith and Deza, Gabriel and Chandrasekaran, Varun and Papernot, Nicolas},
  booktitle={2022 IEEE 7th European symposium on security and privacy (EuroS\&P)},
  pages={303--319},
  year={2022},
  organization={IEEE}
}

@inproceedings{
gao2026illusion,
title={An Illusion of Unlearning? Assessing Machine Unlearning Through Internal Representations},
author={Yichen Gao and Altay Unal and Akshay Rangamani and Zhihui Zhu},
booktitle={The 29th International Conference on Artificial Intelligence and Statistics},
year={2026},
url={https://openreview.net/forum?id=2kwnpGKxUo}
}

@inproceedings{
gandikota2025erasing,
title={Erasing Conceptual Knowledge from Language Models},
author={Rohit Gandikota and Sheridan Feucht and Samuel Marks and David Bau},
booktitle={The Thirty-ninth Annual Conference on Neural Information Processing Systems},
year={2025},
url={https://openreview.net/forum?id=N5V3dlIck9}
}

@inproceedings{belrose2306leace,
  author       = {Nora Belrose and
                  David Schneider{-}Joseph and
                  Shauli Ravfogel and
                  Ryan Cotterell and
                  Edward Raff and
                  Stella Biderman},
  editor       = {Alice Oh and
                  Tristan Naumann and
                  Amir Globerson and
                  Kate Saenko and
                  Moritz Hardt and
                  Sergey Levine},
  title        = {{LEACE:} Perfect linear concept erasure in closed form},
  booktitle    = {Advances in Neural Information Processing Systems 36: Annual Conference
                  on Neural Information Processing Systems 2023, NeurIPS 2023, New Orleans,
                  LA, USA, December 10 - 16, 2023},
  year         = {2023},
  url          = {http://papers.nips.cc/paper\_files/paper/2023/hash/d066d21c619d0a78c5b557fa3291a8f4-Abstract-Conference.html},
  bibsource    = {dblp computer science bibliography, https://dblp.org}
}

@inproceedings{singh2024representation,
  author       = {Shashwat Singh and
                  Shauli Ravfogel and
                  Jonathan Herzig and
                  Roee Aharoni and
                  Ryan Cotterell and
                  Ponnurangam Kumaraguru},
  editor       = {Ruslan Salakhutdinov and
                  Zico Kolter and
                  Katherine A. Heller and
                  Adrian Weller and
                  Nuria Oliver and
                  Jonathan Scarlett and
                  Felix Berkenkamp},
  title        = {Representation Surgery: Theory and Practice of Affine Steering},
  booktitle    = {Forty-first International Conference on Machine Learning, {ICML} 2024,
                  Vienna, Austria, July 21-27, 2024},
  series       = {Proceedings of Machine Learning Research},
  volume       = {235},
  pages        = {45663--45680},
  publisher    = {{PMLR} / OpenReview.net},
  year         = {2024},
  url          = {https://proceedings.mlr.press/v235/singh24d.html},
  bibsource    = {dblp computer science bibliography, https://dblp.org}
}

@article{quirke2025slowing,
  title={Slowing Learning by Erasing Simple Features},
  author={Quirke, Lucia and Belrose, Nora},
  journal={arXiv preprint arXiv:2502.02820},
  year={2025}
}

@inproceedings{feldman2015certifying,
  title={Certifying and removing disparate impact},
  author={Feldman, Michael and Friedler, Sorelle A and Moeller, John and Scheidegger, Carlos and Venkatasubramanian, Suresh},
  booktitle={proceedings of the 21th ACM SIGKDD international conference on knowledge discovery and data mining},
  pages={259--268},
  year={2015}
}

@inproceedings{gordaliza2019obtaining,
  author       = {Paula Gordaliza and
                  Eustasio del Barrio and
                  Fabrice Gamboa and
                  Jean{-}Michel Loubes},
  editor       = {Kamalika Chaudhuri and
                  Ruslan Salakhutdinov},
  title        = {Obtaining Fairness using Optimal Transport Theory},
  booktitle    = {Proceedings of the 36th International Conference on Machine Learning,
                  {ICML} 2019, 9-15 June 2019, Long Beach, California, {USA}},
  series       = {Proceedings of Machine Learning Research},
  volume       = {97},
  pages        = {2357--2365},
  publisher    = {{PMLR}},
  year         = {2019},
  url          = {http://proceedings.mlr.press/v97/gordaliza19a.html},
  bibsource    = {dblp computer science bibliography, https://dblp.org}
}

@inproceedings{bourtoule2021machine,
  title={Machine unlearning},
  author={Bourtoule, Lucas and Chandrasekaran, Varun and Choquette-Choo, Christopher A and Jia, Hengrui and Travers, Adelin and Zhang, Baiwu and Lie, David and Papernot, Nicolas},
  booktitle={2021 IEEE symposium on security and privacy (SP)},
  pages={141--159},
  year={2021},
  organization={IEEE}
}

@inproceedings{cao2015towards,
  title={Towards making systems forget with machine unlearning},
  author={Cao, Yinzhi and Yang, Junfeng},
  booktitle={2015 IEEE symposium on security and privacy},
  pages={463--480},
  year={2015},
  organization={IEEE}
}

@article{eu2016general,
  title={General data protection regulation},
  author={EU, GDPR},
  journal={Official Journal of the European Union},
  year={2016}
}

@article{papyan2020prevalence,
  title={Prevalence of neural collapse during the terminal phase of deep learning training},
  author={Papyan, Vardan and Han, Xiao Y and Donoho, David L},
  journal={Proceedings of the National Academy of Sciences},
  volume={117},
  number={40},
  pages={24652--24663},
  year={2020},
  publisher={National Academy of Sciences}
}

@article{olkin1982distance,
  title={The distance between two random vectors with given dispersion matrices},
  author={Olkin, Ingram and Pukelsheim, Friedrich},
  journal={Linear Algebra and its Applications},
  volume={48},
  pages={257--263},
  year={1982},
  publisher={Elsevier}
}

@article{krizhevsky2009learning,
  title={Learning multiple layers of features from tiny images},
  author={Krizhevsky, Alex and Hinton, Geoffrey and others},
  journal={Technical Report},
  year={2009},
  publisher={Toronto, ON, Canada}
}

@article{le2015tiny,
  title={Tiny imagenet visual recognition challenge},
  author={Le, Yann and Yang, Xuan and others},
  journal={CS 231N},
  volume={7},
  number={7},
  pages={3},
  year={2015}
}

@inproceedings{cao2018vggface2,
  title={Vggface2: A dataset for recognising faces across pose and age},
  author={Cao, Qiong and Shen, Li and Xie, Weidi and Parkhi, Omkar M and Zisserman, Andrew},
  booktitle={2018 13th IEEE international conference on automatic face \& gesture recognition (FG 2018)},
  pages={67--74},
  year={2018},
  organization={IEEE}
}

@inproceedings{zheng2015scalable,
  title={Scalable person re-identification: A benchmark},
  author={Zheng, Liang and Shen, Liyue and Tian, Lu and Wang, Shengjin and Wang, Jingdong and Tian, Qi},
  booktitle={Proceedings of the IEEE international conference on computer vision},
  pages={1116--1124},
  year={2015}
}

@inproceedings{cai2026hcf,
  title={HCF: Hierarchical Cascade Framework for Distributed Multi-Stage Image Compression},
  author={Cai, Junhao and An, Taegun and Jin, Chengjun and Choi, Sung Il and Park, Juhyun and Joo, Changhee},
  booktitle={Proceedings of the AAAI Conference on Artificial Intelligence},
  volume={40},
  number={17},
  pages={14502--14510},
  year={2026}
}

@article{everingham2010pascal,
  title={The pascal visual object classes (voc) challenge},
  author={Everingham, Mark and Van Gool, Luc and Williams, Christopher KI and Winn, John and Zisserman, Andrew},
  journal={International journal of computer vision},
  volume={88},
  number={2},
  pages={303--338},
  year={2010},
  publisher={Springer}
}

@inproceedings{lin2014microsoft,
  title={Microsoft coco: Common objects in context},
  author={Lin, Tsung-Yi and Maire, Michael and Belongie, Serge and Hays, James and Perona, Pietro and Ramanan, Deva and Doll{\'a}r, Piotr and Zitnick, C Lawrence},
  booktitle={European conference on computer vision},
  pages={740--755},
  year={2014},
  organization={Springer}
}

@inproceedings{he2016deep,
  title={Deep residual learning for image recognition},
  author={He, Kaiming and Zhang, Xiangyu and Ren, Shaoqing and Sun, Jian},
  booktitle={Proceedings of the IEEE conference on computer vision and pattern recognition},
  pages={770--778},
  year={2016}
}

@article{oquab2023dinov2,
  author       = {Maxime Oquab and
                  Timoth{\'{e}}e Darcet and
                  Th{\'{e}}o Moutakanni and
                  Huy V. Vo and
                  Marc Szafraniec and
                  Vasil Khalidov and
                  Pierre Fernandez and
                  Daniel Haziza and
                  Francisco Massa and
                  Alaaeldin El{-}Nouby and
                  Mido Assran and
                  Nicolas Ballas and
                  Wojciech Galuba and
                  Russell Howes and
                  Po{-}Yao Huang and
                  Shang{-}Wen Li and
                  Ishan Misra and
                  Michael Rabbat and
                  Vasu Sharma and
                  Gabriel Synnaeve and
                  Hu Xu and
                  Herv{\'{e}} J{\'{e}}gou and
                  Julien Mairal and
                  Patrick Labatut and
                  Armand Joulin and
                  Piotr Bojanowski},
  title        = {DINOv2: Learning Robust Visual Features without Supervision},
  journal      = {Trans. Mach. Learn. Res.},
  volume       = {2024},
  year         = {2024},
  url          = {https://openreview.net/forum?id=a68SUt6zFt},
  bibsource    = {dblp computer science bibliography, https://dblp.org}
}

@inproceedings{radford2021learning,
  title={Learning transferable visual models from natural language supervision},
  author={Radford, Alec and Kim, Jong Wook and Hallacy, Chris and Ramesh, Aditya and Goh, Gabriel and Agarwal, Sandhini and Sastry, Girish and Askell, Amanda and Mishkin, Pamela and Clark, Jack and others},
  booktitle={International conference on machine learning},
  pages={8748--8763},
  year={2021},
  organization={PMLR}
}

@article{cai2026retain,
  title={Retain-Neutral Surrogates for Min-Max Unlearning},
  author={Cai, Junhao and Kim, Dohun and Kim, Dowon and Choi, Sung Il and Jin, Chengjun and Park, Juhyun and Joo, Changhee},
  journal={arXiv preprint arXiv:2605.05871},
  year={2026}
}

@inproceedings{chen2018encoder,
  title={Encoder-decoder with atrous separable convolution for semantic image segmentation},
  author={Chen, Liang-Chieh and Zhu, Yukun and Papandreou, George and Schroff, Florian and Adam, Hartwig},
  booktitle={European conference on computer vision},
  pages={833--851},
  year={2018},
  organization={Springer}
}

@article{ren2015faster,
  title={Faster R-CNN: Towards real-time object detection with region proposal networks},
  author={Ren, Shaoqing and He, Kaiming and Girshick, Ross and Sun, Jian},
  journal={IEEE transactions on pattern analysis and machine intelligence},
  volume={39},
  number={6},
  pages={1137--1149},
  year={2016},
  publisher={IEEE}
}

@article{wang2019dynamic,
  title={Dynamic graph cnn for learning on point clouds},
  author={Wang, Yue and Sun, Yongbin and Liu, Ziwei and Sarma, Sanjay E and Bronstein, Michael M and Solomon, Justin M},
  journal={ACM Transactions on Graphics (TOG)},
  volume={38},
  number={5},
  pages={1--12},
  year={2019},
  publisher={Acm New York, NY, USA}
}

@inproceedings{wu20153d,
  title={3d shapenets: A deep representation for volumetric shapes},
  author={Wu, Zhirong and Song, Shuran and Khosla, Aditya and Yu, Fisher and Zhang, Linguang and Tang, Xiaoou and Xiao, Jianxiong},
  booktitle={Proceedings of the IEEE conference on computer vision and pattern recognition},
  pages={1912--1920},
  year={2015}
}

@article{yi2016scalable,
  title={A scalable active framework for region annotation in 3d shape collections},
  author={Yi, Li and Kim, Vladimir G and Ceylan, Duygu and Shen, I-Chao and Yan, Mengyan and Su, Hao and Lu, Cewu and Huang, Qixing and Sheffer, Alla and Guibas, Leonidas},
  journal={ACM Transactions on Graphics (ToG)},
  volume={35},
  number={6},
  pages={1--12},
  year={2016},
  publisher={ACM New York, NY, USA}
}

@book{hastie2009elements,
  author       = {Trevor Hastie and
                  Robert Tibshirani and
                  Jerome H. Friedman},
  title        = {The Elements of Statistical Learning: Data Mining, Inference, and
                  Prediction, 2nd Edition},
  series       = {Springer Series in Statistics},
  publisher    = {Springer},
  year         = {2009},
  url          = {https://doi.org/10.1007/978-0-387-84858-7},
  doi          = {10.1007/978-0-387-84858-7},
  isbn         = {9780387848570},
  bibsource    = {dblp computer science bibliography, https://dblp.org}
}

@inproceedings{deng2009imagenet,
  title={Imagenet: A large-scale hierarchical image database},
  author={Deng, Jia and Dong, Wei and Socher, Richard and Li, Li-Jia and Li, Kai and Fei-Fei, Li},
  booktitle={2009 IEEE conference on computer vision and pattern recognition},
  pages={248--255},
  year={2009},
  organization={IEEE}
}

@inproceedings{bossard2014food,
  title={Food-101--mining discriminative components with random forests},
  author={Bossard, Lukas and Guillaumin, Matthieu and Van Gool, Luc},
  booktitle={European conference on computer vision},
  pages={446--461},
  year={2014},
  organization={Springer}
}

@inproceedings{stallkamp2011german,
  title={The german traffic sign recognition benchmark: a multi-class classification competition},
  author={Stallkamp, Johannes and Schlipsing, Marc and Salmen, Jan and Igel, Christian},
  booktitle={The 2011 international joint conference on neural networks},
  pages={1453--1460},
  year={2011},
  organization={IEEE}
}

@article{cai2026scope,
  title={SCOPE: Entanglement Frontier Escape for Source-Free Class Unlearning},
  author={Cai, Junhao and Kim, Dohun and Choi, Sung Il and Park, Juhyun and Jin, Chengjun and Kim, Dowon and Joo, Changhee},
  journal={arXiv preprint arXiv:2608.02058},
  year={2026}
}

@techreport{wah2011caltech,
  Title={The Caltech-UCSD Birds-200-2011 Dataset},
  Author={Wah, C. and Branson, S. and Welinder, P. and Perona, P. and Belongie, S.},
  Year={2011},
  Institution = {California Institute of Technology},
  Number = {CNS-TR-2011-001}
}

@article{kim2026data,
  title={Data-Efficient Unlearning-Based Backdoor Detection in Federated Learning},
  author={Kim, Dowon and Cai, Junhao and Choi, Sung Il and Kim, Dohun and Park, Juhyun and Jin, Chengjun and Joo, Changhee},
  journal={arXiv preprint arXiv:2611.13491},
  year={2026}
}

@inproceedings{cui2019class,
  title={Class-balanced loss based on effective number of samples},
  author={Cui, Yin and Jia, Menglin and Lin, Tsung-Yi and Song, Yang and Belongie, Serge},
  booktitle={2019 IEEE/CVF conference on computer vision and pattern recognition (CVPR)},
  pages={9260--9269},
  year={2019},
  organization={IEEE}
}

@article{choi2026spectral,
  title={Spectral Unlearning: Transformer Structure-Preserving Updates for Language Models},
  author={Choi, Sung Il and Cai, Junhao and Kim, Dohun and Joo, Changhee},
  journal={arXiv preprint arXiv:2611.11902},
  year={2026}
}

@inproceedings{liu2021swin,
  title={Swin transformer: Hierarchical vision transformer using shifted windows},
  author={Liu, Ze and Lin, Yutong and Cao, Yue and Hu, Han and Wei, Yixuan and Zhang, Zheng and Lin, Stephen and Guo, Baining},
  booktitle={2021 IEEE/CVF international conference on computer vision (ICCV)},
  pages={9992--10002},
  year={2021},
  organization={IEEE}
}

@inproceedings{dosovitskiy2020image,
  author       = {Alexey Dosovitskiy and
                  Lucas Beyer and
                  Alexander Kolesnikov and
                  Dirk Weissenborn and
                  Xiaohua Zhai and
                  Thomas Unterthiner and
                  Mostafa Dehghani and
                  Matthias Minderer and
                  Georg Heigold and
                  Sylvain Gelly and
                  Jakob Uszkoreit and
                  Neil Houlsby},
  title        = {An Image is Worth 16x16 Words: Transformers for Image Recognition
                  at Scale},
  booktitle    = {9th International Conference on Learning Representations, {ICLR} 2021,
                  Virtual Event, Austria, May 3-7, 2021},
  publisher    = {OpenReview.net},
  year         = {2021},
  url          = {https://openreview.net/forum?id=YicbFdNTTy},
  bibsource    = {dblp computer science bibliography, https://dblp.org}
}

@article{bae2026rxc,
  title={RXC: Runtime Extra Compression for Time-Critical Content Delivery},
  author={Bae, Jeongmin and Kang, Harim and Kim, Taehun and An, Taegun and Lee, Seunghyun and Cai, Junhao and Joo, Changhee and Lee, Kyunghan},
  journal={IEEE Communications Magazine},
  year={2026},
  publisher={IEEE}
}

@article{radic2026forget,
  title={Why Forget-Only Unlearning Needs Memorization},
  author={Radi{\'c}, Luka and Singhal, Vikrant and Sanyal, Amartya},
  journal={arXiv preprint arXiv:2610.10519},
  year={2026}
}

@article{delon2020wasserstein,
  title={A Wasserstein-type distance in the space of Gaussian mixture models},
  author={Delon, Julie and Desolneux, Agnes},
  journal={SIAM Journal on Imaging Sciences},
  volume={13},
  number={2},
  pages={936--970},
  year={2020},
  publisher={SIAM}
}

@inproceedings{deng2019arcface,
  title={Arcface: Additive angular margin loss for deep face recognition},
  author={Deng, Jiankang and Guo, Jia and Xue, Niannan and Zafeiriou, Stefanos},
  booktitle={Proceedings of the IEEE/CVF conference on computer vision and pattern recognition},
  pages={4690--4699},
  year={2019}
}

@inproceedings{cai2024toward,
  title={Toward scalable and efficient visual data transmission in 6G networks},
  author={Cai, Junhao and An, Taegun and Joo, Changhee},
  booktitle={2024 15th International Conference on Information and Communication Technology Convergence (ICTC)},
  pages={545--547},
  year={2024},
  organization={IEEE}
}
}
\end{document}